%% file: Mu-Edit.tex
\documentclass{article}

\usepackage{amsmath} 
\usepackage{amsthm}

\usepackage{graphicx} 
\usepackage{xcolor} 
\definecolor{blue}{RGB}{191,223,255} 
\usepackage{colortbl}
\usepackage[main,final]{neurips_2026}
\usepackage{amsfonts}
\usepackage{amsmath,amssymb,mathtools,amsthm}
\usepackage{hyperref}
\usepackage{algorithm}
\usepackage{algorithmic}
\usepackage{multirow}

\usepackage[utf8]{inputenc} 
\usepackage[T1]{fontenc}    
\usepackage{hyperref}       
\usepackage{url}            
\usepackage{booktabs}       
\usepackage{amsfonts}       
\usepackage{nicefrac}       
\usepackage{microtype}      
\usepackage{xcolor}         
\usepackage{amsthm}  

\newtheorem{lemma}{Lemma}
\title{MuEdit: An efficient multi-task editing method towards inter-domain knowledge conflicts}

\begin{document}

\author{%
  \begin{tabular}{@{}ccccc@{}}
    \textbf{Songshi Liang}$^{1}$ &
    \textbf{Ting-En Lin}$^{2}$ &
    \textbf{Zixuan Huang}$^{3}$ &
    \textbf{Yuchuan Wu}$^{2}$ &
    \textbf{Yongbin Li}$^{2}$
  \end{tabular}
  \\[0.4em]
  \begin{tabular}{@{}cc@{}}
    \textbf{Zihe Wang}$^{1\dagger}$ &
    \textbf{Rui Yan}$^{4,5\dagger}$
  \end{tabular}
  \\[0.6em]
  {\normalfont
    $^{1}$Gaoling School of Artificial Intelligence,
    Renmin University of China
  }
  \\
  {\normalfont
    $^{2}$Alibaba Group
    \quad
    $^{3}$Beihang University
  }
  \\
  {\normalfont
    $^{4}$School of Artificial Intelligence, Wuhan University
  }
  \\
  {\normalfont
    $^{5}$School of Computer Science, Wuhan University
  }
  \\[0.4em]
  {\normalfont\small
    \texttt{\{lssgl2024,wang.zihe\}@ruc.edu.cn}
  }
  \\
  {\normalfont\small
    \texttt{\{ting-en.lte,shengxiu.wyc,shuide.lyb\}@alibaba-inc.com}
  }
  \\
  {\normalfont\small
    \texttt{huang\_zx@buaa.edu.cn}
    \quad
    \texttt{rui.yan@whu.edu.cn}
  }
}

\maketitle
\begingroup
  \renewcommand{\thefootnote}{\fnsymbol{footnote}}
  \footnotetext[2]{Corresponding authors.}
\endgroup

\begin{abstract}
  Large language models (LLMs) require frequent knowledge updates across heterogeneous domains, but existing editing methods fail when applied sequentially to multi-task settings. We trace this collapse to \textit{inter-domain null-space misalignment}: domain-specific preservation representations can leave a small common subspace for admissible updates. We propose the \textbf{Conflict Index} to quantify this geometric interference, and introduce \textbf{Mu-Edit}, which mitigates multi-task conflicts via (1) \textbf{conflict-aware ordering} to minimize cumulative interference, and (2) \textbf{dynamic low-rank approximation} to expand the null space when ordering alone is insufficient. Experiments across five functional domains and three model backbones show improved multi-task editing performance while retaining general capabilities. Additional seven-task experiments support the benefits of conflict-aware ordering, and greedy ordering achieves performance close to offline best-order search with substantially lower ordering cost.
\end{abstract}

\section{Introduction}

Large language models (LLMs) require updates as information and application requirements change~\citep{Huang2026RealTimeAR,zhang2026heterogeneous}. Model editing aims to incorporate these updates without retraining the entire model or disrupting unrelated capabilities~\citep{meng2022locating,meng2022mass,ni2023forgetting}. In deployment, updates may span factual knowledge, multilingual knowledge, safety-related behavior, question answering, and sentiment classification~\citep{huang2025adaptive, huang2026deferred,huang2026does}. Maintaining the benefits of earlier edits across such heterogeneous domains is a central challenge for model editing.

Existing editing methods primarily address \textit{sequential editing}, where a model undergoes successive updates within a \textbf{single task domain} (e.g., correcting thousands of factual associations in commonsense knowledge)~\citep{meng2022mass, mitchell2021fast}.  AlphaEdit~\citep{fang2024alphaedit} uses null-space constraints derived from preserved knowledge representations, whereas PRUNE~\citep{ma2024perturbation} controls accumulated editing perturbations through their singular values. These methods address degradation under repeated editing, but their effectiveness within a domain does not establish robustness across heterogeneous domains.

In this work, we formalize and investigate \textit{multi-task model editing}, \textbf{in which a model is edited sequentially across functionally distinguishable domains without resetting its parameters between tasks}. Each domain is characterized by its own reference data and task-specific knowledge geometry. The editing process proceeds as a chain of parameter updates, where the model is edited on one domain after another, culminating in the final state that carries knowledge from all domains.

\begin{figure*}[t]  
    \centering
    \includegraphics[width=0.93\textwidth]{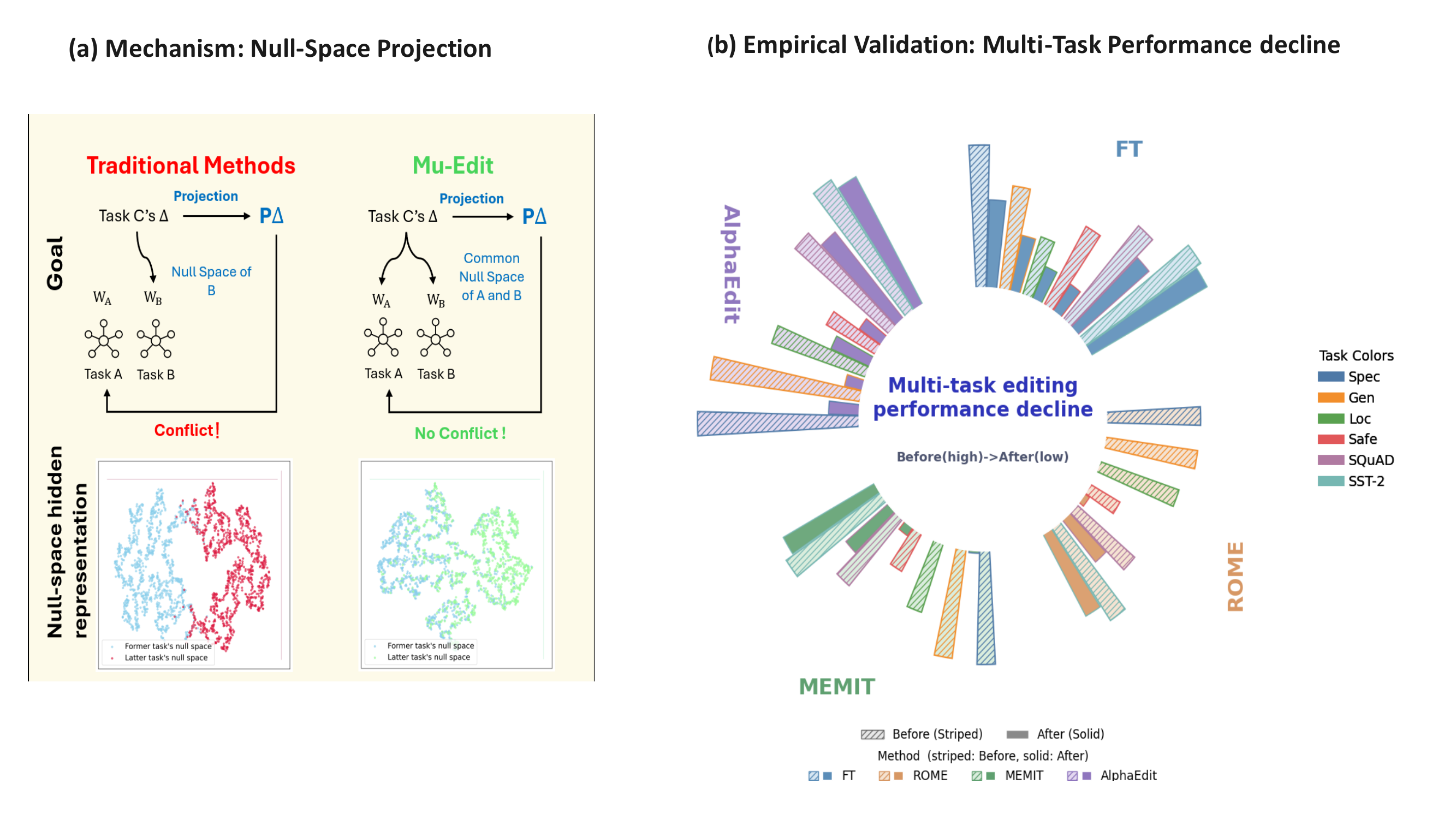}
    \caption{Mechanism and motivation of our Mu-Edit, specifically: (a) Visualization of Mu-Edit's null space projection mechanism. (b) Empirical validation results of existing methods' performance decline after editing sequentially across different tasks.}
    \label{fig:backbone}
\end{figure*}

In our definition, the core goal of multi-task editing is to update knowledge across different functional domains sequentially while maintaining performance on previously edited domains. These domains need not be independent or have disjoint semantic content.
As shown in Fig.~\ref{fig:backbone}(b), current methods including null space projection methods such as AlphaEdit~\citep{fang2024alphaedit}, suffer from severe performance degradation under multi-task editing. They frequently forget knowledge injected in earlier tasks as subsequent domains are edited, and the degradation becomes more pronounced as the number of tasks accumulates. 
Notably, this collapse is \textit{task-pair-dependent}: some transitions induce mild forgetting, whereas others cause catastrophic interference, motivating an analysis of inter-domain geometric compatibility.

Our analysis identifies \emph{inter-domain null-space misalignment} as a geometric constraint on multi-task editing. Each domain induces a preservation matrix with its own representation subspace. An update that preserves the outputs associated with these representations must satisfy their joint null-space constraints. Although each domain may individually admit many update directions, their intersection can be much smaller. This restricts the directions available to fit a new edit while preserving existing representations. Building on this insight, we propose to mitigate multi-task interference by explicitly measuring and maximizing the common null space shared among tasks. 
To operationalize this idea, we introduce a new concept called the \textbf{Conflict Index}, which quantifies the geometric overlap between task-specific null spaces. Using this index, we develop a novel framework called \textbf{Mu-Edit}, which incorporates two complementary strategies to resolve multi-task conflicts. First, conflict-aware ordering minimizes adjacent pairwise conflicts as a surrogate for editing interference. Second, adaptive low-rank approximation relaxes selected preservation constraints to expand the admissible update space. The compressed representation is retained for subsequent computations.
For scenarios with increasing or unpredictable task numbers, we further develop a \textbf{Greedy Search} strategy to handle the excessive computational complexity of best-order search and to enable editing under incremental task arrival. Our main contributions are three-fold:
\begin{enumerate}
\item To the best of our knowledge, this work is \textbf{the first to investigate multi-task knowledge editing from the perspective of null-space conflicts}. We propose the concept of \textbf{Conflict Index} based on the null-space properties of task-specific matrices, enabling effective quantification of the conflict degree between different tasks.
\item Building on this Conflict Index, we develop \textbf{Mu-Edit}, combining conflict-aware ordering with adaptive rank reduction to balance edit fitting and representation preservation.
\item We evaluate Mu-Edit across five domains and three backbones, with additional seven-task, reference-data, and computational-cost analyses. The results support improved final editing performance and show that greedy ordering approaches offline best-order performance at lower ordering cost.
\end{enumerate}

\section{Preliminary}
\subsection{Knowledge Storage and Editing in LLMs}
For a specific token at layer $i$, let
$\tilde{h}^{i} \in \mathbb{R}^{d}$ denote the hidden
representation supplied to the MLP. For a standard
two-layer MLP, its output representation $h^i$ is
computed as
\begin{equation}
    h^i = W_2^i \sigma(W_1^i \tilde{h}^i),
    \nonumber
\end{equation}
where $W_1^i \in \mathbb{R}^{d_i \times d}$ projects
the input into the intermediate space,
$\sigma(\cdot)$ denotes the activation function, and
$W_2^i \in \mathbb{R}^{d_o \times d_i}$ maps the
intermediate activation to the output space.
Here, $h^i$ denotes the MLP output rather than the
complete hidden state after the residual connection.

Following prior work~\citep{geva-etal-2021-transformer,
meng2022locating,meng2022mass}, the MLP can be
interpreted as a key--value associative memory.
The intermediate activation serves as a key,
and the output projection maps it to a value:
\begin{equation}
    k^i = \sigma(W_1^i \tilde{h}^i),
    \qquad
    v^i = W_2^i k^i = h^i.
    \nonumber
\end{equation}
The key captures features of the input context,
while the value contributes information to subsequent
model computation. For gated MLPs, the key is produced
by the corresponding gated transformation, while the
output mapping $v^i=W_2^i k^i$ remains applicable.
Below, we omit the layer index and denote the
output-projection matrix by $W$.
\subsection{Traditional model-editing methods}
Model editing aims to update knowledge stored in LLMs
through individual edits or multiple edits performed
jointly or sequentially. Within the locate-then-edit
paradigm, we first identify the parameters to modify
and then add a perturbation $\Delta$ to the selected
weight matrix $W$, yielding $\widehat{W}=W+\Delta$. For factual knowledge editing, a knowledge association
can be represented as a triplet $(s,r,o)$, where $s$,
$r$, and $o$ denote the subject, relation, and object, respectively. Under the associative-memory view,
the key $k$ encodes the subject--relation context
$(s,r)$, while the target value $v$ represents
information associated with the desired object $o$.
An edit therefore aims to map the corresponding key
to its target value while preserving unrelated
key--value associations.

The weight matrix $W \in \mathbb{R}^{d_o \times d_i}$
maps keys in the FFN's intermediate space to values
in its output space. For $u$ existing associations
to be preserved, we define
\begin{equation}
K_0 = [k_1,\ldots,k_u] \in \mathbb{R}^{d_i \times u},
\quad
V_0 = [v_1,\ldots,v_u] \in \mathbb{R}^{d_o \times u}.
\nonumber
\end{equation}
where each column corresponds to one association,
and $v_j=Wk_j$ is its pre-edit output.
Thus, $V_0=WK_0$ represents the outputs to be
preserved. A new association to be introduced is
denoted by an edit key $k_e$ and its target value
$v_e$.

ROME~\citep{meng2022locating} introduces a new key--value
association $(k_e,v_e)$ while minimizing changes to the
outputs associated with the preserved keys:
\begin{equation}
    \widehat{W}
    = \arg\min_{\widehat{W}:\,\widehat{W}k_e=v_e}
    \underbrace{
        \|\widehat{W}K_0-WK_0\|_F^2
    }_{\text{preservation}}.
    \nonumber
\end{equation}
The equality constraint enforces the new association,
while the preservation term discourages changes to
existing associations.

Whereas ROME targets a single association per update,
MEMIT~\citep{meng2022mass} supports batched editing
through a relaxed objective:
\begin{equation}
    \widehat{W}
    = \arg\min_{\widehat{W}}
    \underbrace{
        \lambda\|\widehat{W}K_0-WK_0\|_F^2
    }_{\text{preservation}}
    +
    \underbrace{
        \|\widehat{W}K_E-V_E\|_F^2
    }_{\text{memorization}},
    \label{eq:memit_objective}
\end{equation}
where $K_E=[k_1^e,\ldots,k_E^e]$ and
$V_E=[v_1^e,\ldots,v_E^e]$ contain the keys and target
values of $E$ editing associations as columns.
The coefficient $\lambda$ balances preservation
against memorization. Unlike ROME's equality constraint,
the memorization term fits the new associations by
minimizing their squared reconstruction error.

\vspace{-0.05cm}
\subsection{Sequential-optimized editing methods}
PRUNE~\citep{ma2024perturbation} and
AlphaEdit~\citep{fang2024alphaedit} address model
degradation under repeated editing through different
mechanisms. PRUNE restrains editing-induced perturbations
to mitigate their accumulated effects on the model's
existing knowledge and general capabilities.
AlphaEdit instead uses null-space projection to protect
preserved knowledge representations.

To explain the AlphaEdit's mechanism, recall that $K_0$ contains
the keys of the knowledge to be preserved.
Its left null space is
\begin{equation}
    \mathcal{N}(K_0^\top)
    = \{x \in \mathbb{R}^{d_i}: K_0^\top x=0\}.
    \nonumber
\end{equation}
An update whose rows lie in this space leaves the
outputs associated with $K_0$ unchanged.
Let $P$ denote the orthogonal projector onto
$\mathcal{N}(K_0^\top)$, satisfying $PK_0=0$.
AlphaEdit parameterizes the applied update as
$\Delta P$, yielding
\begin{equation}
    \widehat{W}K_0
    = (W+\Delta P)K_0
    = WK_0
    = V_0.
    \nonumber
\end{equation}
Thus, the projected update preserves the existing
key--value associations represented by $K_0$ at the
edited linear map while allowing new associations
to be fitted through the available update directions.
Here, the projection is constructed from preserved
knowledge representations, rather than solely from
knowledge injected by previous edits.

\section{Definition of Multi-task Editing}
In multi-task editing, we aim to update knowledge across
$N$ functionally distinguishable domains
$\{I_1,I_2,\ldots,I_N\}$ while preserving performance
on previously edited domains.
The editing process proceeds sequentially without
resetting model parameters: starting from initial
parameters $\mathbf{W}_0$, we obtain $\mathbf{W}_n$
after editing on task $I_n$, and the final parameters
are $\mathbf{W}^{*}=\mathbf{W}_N$. Each domain $I_n$ is associated with its own reference
data distribution and knowledge matrix $K_n$.
Compared with sequential editing within a single
domain, multi-task editing introduces heterogeneity
in the representations to be preserved, rather than
simply increasing the number of edits.
Different domains may induce distinct dominant
subspaces in their knowledge matrices, making update
directions that preserve one domain incompatible with
the preservation constraints of another.

\subsection{Geometric Analysis of Multi-task null-space overlap and measurement}
Null-space projection methods such as
AlphaEdit~\citep{fang2024alphaedit} preserve existing
knowledge representations by constraining parameter
updates to the corresponding null space.
In multi-task editing, each domain contributes a
knowledge matrix $K_i$ representing the associations
to be preserved. An update $\Delta$ that preserves
the outputs associated with every such matrix must
satisfy $\Delta K_i=0$ for each domain.
Consequently, each row of $\Delta$ must lie in the
intersection of the corresponding left null spaces.
We refer to this intersection as the \textit{common
null space}: it contains the update directions that
simultaneously preserve the representations of the
domains under consideration.
To characterize how combining preservation constraints
affects the available update space, we first consider
two domains with knowledge matrices $K_i$ and $K_j$.
Their common null space satisfies
\begin{equation}
\begin{aligned}
    \mathcal{N}([K_i,K_j]^\top)
    &= \mathcal{N}(K_i^\top)
    \cap \mathcal{N}(K_j^\top), \\
    \dim\mathcal{N}([K_i,K_j]^\top)
    &\leq
    \min\{\dim\mathcal{N}(K_i^\top),
           \dim\mathcal{N}(K_j^\top)\}.
\end{aligned}
\end{equation}
The proof is given in Appendix Lemma~\ref{lemma:nullspace}.
Although each domain may individually admit a large
space of update directions, the intersection can be
much smaller when their preservation subspaces differ.
A smaller common null space restricts
the directions available to fit new associations
while preserving existing representations.
If the new edit cannot be accommodated within this
space, enforcing the preservation constraints limits
edit fitting, whereas relaxing them can alter
previously preserved outputs.
We refer to this geometric bottleneck as
\textit{inter-domain null-space misalignment}.

The analysis above suggests that inter-domain conflicts
can be characterized by how much of the available update
space remains when two preservation constraints are
imposed jointly. We therefore compare the dimension of
the common null space with those of the individual
null spaces.

For tasks $i$ and $j$ at layer $l$, the individual
update spaces are $\mathcal{N}(K_{i,l}^{\top})$ and
$\mathcal{N}(K_{j,l}^{\top})$, and their common null
space is $\mathcal{N}([K_{i,l},K_{j,l}]^{\top})$.
Here, $[K_{i,l},K_{j,l}]$ denotes horizontal
concatenation.
For positive individual null-space dimensions,
we define the layer-wise conflict score as
\begin{equation}
    \mathcal{C}(K_{i,l},K_{j,l})
    =
    1-
    \frac{
        \dim\mathcal{N}([K_{i,l},K_{j,l}]^\top)
    }{
        \min\{
            \dim\mathcal{N}(K_{i,l}^\top),
            \dim\mathcal{N}(K_{j,l}^\top)
        \}
    }.
\end{equation}
The ratio measures the fraction of the
smaller null space that remains available under both
constraints; subtracting it from one gives the
relative loss of admissible update directions.
Averaging over the $L$ edited layers yields the
task-level \textbf{Conflict Index}:
\begin{equation}
    \mathcal{C}(K_i,K_j)
    =
    \frac{1}{L}\sum_{l=1}^{L}
    \mathcal{C}(K_{i,l},K_{j,l}).
    \label{eq:conflict_task}
\end{equation}
By construction, the index lies in $[0,1]$.
At each layer, a score of zero means that the smaller
null space is entirely contained in the larger one,
whereas a score of one means that no nonzero update
direction satisfies both preservation constraints.
The Conflict Index provides a geometric measure of
compatibility between task-specific preservation
constraints. We use it to guide task ordering and
adaptive low-rank approximation, and empirically
evaluate its robustness to limited and mixed reference
data in our experiments.

\subsection{Reducing Cumulative Interference Through Task Ordering}

The Conflict Index measures pairwise geometric
incompatibility between preservation constraints.
To use this information for multi-task editing,
we must also consider how these constraints are
encountered along an editing sequence. Ideally, an update that preserves the representations
of all preceding tasks must lie, row-wise, in the
intersection of their null spaces. Adding a task can reduce or leave unchanged the feasible space. For fixed task matrices, the final
intersection is independent of task order.
However, different orders expose intermediate
editing steps to different sets of preservation
constraints and produce different parameter-update
trajectories. This motivates selecting an order that reduces
conflicts between consecutive tasks. We use adjacent
pairwise Conflict Indices as a tractable surrogate
for editing interference, rather than an exact
measure of cumulative forgetting.
The information available about future tasks
determines how this ordering strategy can be
applied. We therefore consider two complementary deployment scenarios.

\paragraph{Offline setting: Best Order.}
When all $N$ tasks and their reference data are
available in advance, we select an editing order
that minimizes the sum of adjacent pairwise
Conflict Indices:
\begin{equation}
    \pi^\star \in \arg\min_{\pi \in S_N}
    \sum_{n=1}^{N-1}
    \mathcal{C}(K_{\pi(n)},K_{\pi(n+1)}),
    \label{eq:best_order}
\end{equation}
where $S_N$ denotes the set of all permutations
of the $N$ tasks.
We refer to this strategy as Best Order (BO).
It provides the optimal ordering under the
pairwise surrogate objective and serves as an
offline reference for evaluating cheaper ordering
strategies. Our implementation searches all $N!$
permutations, making exhaustive BO practical
only for small task sets.

\paragraph{Online setting: Greedy Order.}
When tasks arrive incrementally and future tasks
are unknown, we use Greedy Order (GO) to maintain
a conflict-aware task sequence without replaying
previous edits.
When a new task $t_{\mathrm{new}}$ arrives,
GO computes its Conflict Index with existing
tasks and selects the insertion position that
minimizes the sum of conflicts with its immediate
neighbors. The relative order of existing tasks
remains unchanged.

The selected position determines the local
protection window for the new edit.
A window of size $k$ consists of the current
task and the $k-1$ tasks immediately preceding
it in the maintained sequence.
Let $\mathcal{W}_{k-1}(t_{\mathrm{new}})$
denote these preceding tasks.
We construct the projection matrix $P$ onto
their common null space:
\begin{equation}
    \bigcap_{j \in \mathcal{W}_{k-1}(t_{\mathrm{new}})}
    \mathcal{N}(K_j^\top).
    \nonumber
\end{equation}
The projected update is applied to the current
model without re-executing previous edits.
Thus, the logical insertion position affects
the actual parameter update by determining
which preceding task representations are
protected. This local projection avoids imposing the
constraints of all historical tasks at every
editing step. We use $k=3$ by default.
GO requires $O(N)$ pairwise conflict evaluations
per insertion into a sequence of $N$ existing
tasks. Algorithm~\ref{alg:greedy_insertion}
provides the insertion procedure.

\subsection{Expanding the Common Null Space When Ordering Is Insufficient}

Task ordering reduces interference by avoiding high-conflict adjacencies, but it cannot eliminate conflict entirely when the underlying null-space intersection is inherently small.
When the Conflict Index between two tasks exceeds
a threshold $\mu$, we treat their preservation
constraints as highly conflicting and apply
\textit{conflict-guided low-rank approximation}.
By discarding selected singular directions from
a preservation matrix, we relax its constraints
and expand the space available for subsequent updates.

\paragraph{Dynamic rank reduction.}
Let $K$ be the knowledge matrix selected for
compression, with rank $R$ and singular values
$\sigma_1 \geq \cdots \geq \sigma_R > 0$.
We retain its leading $d_\sigma$ singular components:
\begin{equation}
    \widetilde{K}
    =
    \sum_{r=1}^{d_\sigma}
    \sigma_r u_r v_r^\top,
    \label{eq:svd}
\end{equation}
where $u_r$ and $v_r$ are the corresponding left
and right singular vectors.
Since
$\mathcal{N}(K^\top)
\subseteq \mathcal{N}(\widetilde{K}^\top)$,
this approximation expands the selected matrix's
left null space whenever $d_\sigma<R$.
We further determine the retained rank from the amount
by which the pairwise conflict exceeds $\mu$:
\begin{equation}
    d_\sigma =
    \begin{cases}
        R,
        & \mathcal{C}(K_i,K_j)\leq\mu, \\[2pt]
        \operatorname{round}\!\left(
            R\left[1-\alpha
            \bigl(\mathcal{C}(K_i,K_j)-\mu\bigr)\right]
        \right),
        & \mathcal{C}(K_i,K_j)>\mu,
    \end{cases}
    \label{eq:rank_reduction}
\end{equation}
where $R$ is the rank before compression,
$\alpha$ controls the compression strength,
and $\operatorname{round}(\cdot)$ rounds to
the nearest integer.
No reduction is applied below the threshold;
above it, larger conflict excess leads to a
smaller retained rank, subject to rounding.
With the default $\alpha=1$, a nonnegative
threshold $\mu$, and
$\mathcal{C}(K_i,K_j)\in[0,1]$,
the retained rank lies in $[0,R]$.
The rank-reduction rule determines how much to
compress, but applying it also requires deciding
when compression is needed and which task matrix
to modify. We make these decisions using each
task's conflict distribution and its aggregate
conflict with the other available tasks.

\paragraph{Adaptive thresholds and compression target selection.} 
We first determine the pairwise threshold using
the mean and standard deviation of each task's
conflict scores:
\begin{equation}
    \overline{C}_i
    = \frac{1}{N}\sum_{j=1}^{N}
      \mathcal{C}(K_i,K_j),
    \qquad
    \mu_{ij}
    = \min\left\{
        \overline{C}_i+t\sigma_{C_i},
        \overline{C}_j+t\sigma_{C_j}
      \right\},
    \label{eq:adaptive_threshold}
\end{equation}
where $\sigma_{C_i}$ denotes the standard
deviation of task $i$'s conflict scores and
$t$ controls the threshold sensitivity
(default $0.9$).
Here, $N$ denotes the number of currently
available tasks.
Taking the smaller of the two task-specific
thresholds triggers compression when the
pairwise conflict exceeds either task's
threshold. We use $\mu_{ij}$ in place of
$\mu$ in Equation~\ref{eq:rank_reduction}.

When compression is triggered, either $K_i$
or $K_j$ can be approximated to relax the
pair's preservation constraints.
To choose between them, we compare their
aggregate conflicts with the other available
tasks and select the more conflicting matrix:
\begin{equation}
    q =
    \begin{cases}
        i,
        & \displaystyle
          \sum_{r\neq i}\mathcal{C}(K_i,K_r)
          >
          \sum_{r\neq j}\mathcal{C}(K_j,K_r), \\[4pt]
        j,
        & \text{otherwise}.
    \end{cases}
    \label{eq:select_compress}
\end{equation}
This rule allocates compression to the task
with the greater overall conflict burden,
while leaving the other task's representation
unchanged at this step. We then compute the rank-$d_\sigma$
approximation $\widetilde{K}_q$ using
Equation~\ref{eq:svd} and replace the maintained
matrix. This replacement persists across subsequent
editing steps: later conflict computations
and local protection-window projections use
the updated representation.

\subsection{Closed-Form Parameter Update with Pseudo-Inverse}

After selecting the task position and adapting the
preservation matrices, we compute the update to the
current model parameters $W$.
We use a three-task local window consisting of the
current task and its two immediate predecessors
in the maintained sequence.
For notational convenience, we index these tasks
by $n+1$, $n$, and $n-1$, respectively.

Let $K_{n-1}$ and $K_n$ denote the maintained
preservation matrices after any low-rank adaptation.
We construct the orthogonal projector $P$ onto
$\mathcal{N}([K_{n-1},K_n]^\top)$.
Let $K_{E,n+1}$ and $V_{E,n+1}$ contain the current
task's edit keys and target values.
The applied parameter update $\Delta$ satisfies
$\Delta=\Delta P$ and minimizes
\begin{equation}
\begin{aligned}
    \Delta^\star
    = \arg\min_{\Delta:\,\Delta=\Delta P}
    \;& \|(W+\Delta)K_{E,n+1}-V_{E,n+1}\|_F^2
       + \|\Delta\|_F^2 \\
    &+ \|\Delta K_{n-1}\|_F^2
       + \|\Delta K_n\|_F^2.
\end{aligned}
\label{eq:objective}
\end{equation}
The first term fits the new associations, while
the second regularizes the update magnitude.
The last two terms measure changes to the
representations of the preceding tasks.
Under exact projection, these two terms vanish
because $PK_{n-1}=PK_n=0$.

Define the residual and quadratic coefficient as
\begin{equation}
\begin{aligned}
    E &= V_{E,n+1}-WK_{E,n+1}, \\
    M &= K_{E,n+1}K_{E,n+1}^\top + I
       + K_{n-1}K_{n-1}^\top + K_nK_n^\top.
\end{aligned}
\nonumber
\end{equation}
The projected quadratic coefficient $PMP$ is
symmetric positive semidefinite.
The identity term in $M$ makes it positive
definite on the range of $P$, although it may
remain singular in the full parameter space.
Using the Moore--Penrose pseudo-inverse gives
\begin{equation}
    \Delta^\star
    =
    E K_{E,n+1}^\top P(PMP)^\dagger,
    \label{eq:closed_form_update}
\end{equation}
where $(\cdot)^\dagger$ denotes the pseudo-inverse.
We then apply the update as
$W \leftarrow W+\Delta^\star$.

This formulation accommodates rank-deficient
projections and preserves the retained
representations within the selected local window.
We use a window size of three to balance editing
performance and computational cost.
Appendix Table~\ref{number of tasks}
compares this setting with windows of four
and five tasks.
\section{Experiments}

\subsection{Experimental Setup}

\noindent\textbf{Datasets.}
We evaluate on five functional domains following
EasyEdit2~\citep{xu2025easyedit2}: commonsense
(ZsRE~\citep{wang2023easyedit},
CounterFact~\citep{meng2022locating}),
safety (SafeEdit~\citep{wang2024detoxifying}),
question answering and reasoning
(SQuAD~\citep{rajpurkar2016squad}, GSM8K),
sentiment (SST-2~\citep{socher2013recursive}),
and multilingual knowledge
(CKnowEdit~\citep{xu2025easyedit2}).
Our main experiments use ZsRE, CKnowEdit,
SafeEdit, SQuAD, and SST-2.
By default, we use 500 examples per task for
editing and 10,000 reference examples per task
to construct the preservation matrix $K$.
Experiments with alternative task combinations are provided in the Appendix.

\noindent\textbf{Metrics and baselines.}
For ZsRE, CounterFact, and CKnowEdit, we report
Specificity, Generalization, and Locality.
For SafeEdit, we report Harmful Rate;
for SQuAD, GSM8K, and SST-2, we report Accuracy.
Definitions for ZsRE and CKnowEdit metrics are provided in Appendix Section~\ref{Matrix_description}.
All metrics measure final performance after
completing the entire editing sequence. We compare Mu-Edit with fine-tuning (FT),
ROME~\citep{meng2022locating},
MEMIT~\citep{meng2022mass},
SERAC~\citep{mitchell2022memory},
PRUNE~\citep{ma2024perturbation},
AlphaEdit~\citep{fang2024alphaedit},
and AnyEdit~\citep{jiang2025anyedit}.
The main experiments use Llama3-8B~\citep{dubey2024llama}, with
Mu-Edit applied to layers 4--8.
Method-specific implementation settings are
provided in the Appendix.
Results on Qwen2.5-7B~\citep{yang2024qwen25} and GPT2-xl~\citep{radford2019language} are reported
in Tables~\ref{tab:final_qwen25}
and~\ref{tab:final_gpt2_xl}, respectively.

\noindent\textbf{Reference data.}
We construct preservation matrices from external
reference inputs without using task labels or
evaluation answers.
Mu-Edit and AlphaEdit share the same reference
samples and preprocessing.
Exact normalized-text duplicates with editing
and evaluation data are removed and replaced.
Reference sources and construction details
are provided in the Appendix.

\subsection{Main Results and Ablation Studies}

\begin{table*}[t]
\caption{Final performance after multi-task editing on Llama3-8B under Best Order(BO) and Default Order(DO) and ablations. Bold: best. *: significant improvement over AnyEdit ($p<0.05$).}
\label{tab:main_bo}
\centering
\resizebox{0.95\linewidth}{!}{
\begin{tabular}{l|ccc|ccc|c|cc}
\toprule
 & \multicolumn{3}{c|}{ZsRE} & \multicolumn{3}{c|}{CKnowEdit} & SafeEdit & SQuAD & SST-2 \\
\cmidrule(lr){2-4}\cmidrule(lr){5-7}\cmidrule(lr){8-8}\cmidrule(lr){9-9}\cmidrule(lr){10-10}
Method & Spec$\uparrow$ & Gen$\uparrow$ & Loc$\uparrow$ & Spec$\uparrow$ & Gen$\uparrow$ & Loc$\uparrow$ & Harm$\uparrow$ & Acc$\uparrow$ & Acc$\uparrow$ \\
\midrule
AlphaEdit(DO) & 0.3009 & 0.2351 & 0.3608 & 0.2873 & 0.2512 & 0.3303 & 0.2778 & 0.6984 & 0.8620 \\
AnyEdit(DO) & 0.3954 & 0.3061 & 0.4021 & 0.3741 & 0.3202 & 0.3495 & 0.3004 & 0.7028 & 0.8643 \\
\midrule
Mu-Edit(DO) & 0.8197 & 0.7995 & 0.5684 & 0.7925 & 0.7606 & 0.5642 & 0.3648 & 0.7252 & 0.8694 \\
\midrule
FT(BO)& 0.2861 & 0.2356 & 0.2575 & 0.2559 & 0.2216 & 0.2083 & 0.2210 & 0.4446 & 0.6290 \\
ROME(BO) & 0.3452 & 0.3036 & 0.2418 & 0.3021 & 0.2746 & 0.2327 & 0.2449 & 0.4808 & 0.7073 \\
MEMIT(BO) & 0.2861 & 0.2356 & 0.2575 & 0.2546 & 0.2227 & 0.2083 & 0.2210 & 0.4446 & 0.6290 \\
SERAC(BO) & 0.3827 & 0.3614 & 0.3116 & 0.5142 & 0.4336 & 0.3758 & 0.3009 & 0.6698 & 0.8596 \\
PRUNE(BO) & 0.4892 & 0.4784 & 0.4253 & 0.6819 & 0.6622 & 0.4613 & 0.3446 & 0.6887 & 0.8726 \\
AlphaEdit(BO) & 0.5506 & 0.5101 & 0.4470 & 0.6624 & 0.6313 & 0.4456 & 0.3396 & 0.7016 & 0.8803 \\
AnyEdit(BO) & 0.7098 & 0.6799 & 0.5346 & 0.6882 & 0.6574 & 0.5158 & 0.3691 & 0.7166 & 0.8798 \\
\midrule
Mu-Edit$^-$(BO) & 0.8464 & 0.8077 & 0.6276 & 0.8229 & 0.7819 & 0.6052 & 0.3718 & 0.6962 & 0.8817 \\
Mu-Edit(left)(BO) & 0.8107 & 0.7753 & 0.5889 & 0.7942 & 0.7551 & 0.5686 & 0.3677 & 0.6835 & 0.8789 \\
Mu-Edit(right)(BO) & 0.8556 & 0.8210 & 0.6195 & 0.8342 & 0.8066 & 0.5945 & 0.3895 & 0.7183 & 0.8815 \\
Mu-Edit(BO) & \textbf{0.8845}* & \textbf{0.8452}* & \textbf{0.6321}* & \textbf{0.8505}* & \textbf{0.8258}* & \textbf{0.6144}* & \textbf{0.4043}* & \textbf{0.7479}* & \textbf{0.8838}* \\
\bottomrule
\end{tabular}
}
\end{table*}

\begin{table*}[t]
\caption{Final performance under Greedy Order (GO) and ablations.}
\label{tab:main_go}
\centering
\resizebox{0.95\linewidth}{!}{
\begin{tabular}{l|ccc|ccc|c|cc}
\toprule
 & \multicolumn{3}{c|}{ZsRE} & \multicolumn{3}{c|}{CKnowEdit} & SafeEdit & SQuAD & SST-2 \\
\cmidrule(lr){2-4}\cmidrule(lr){5-7}\cmidrule(lr){8-8}\cmidrule(lr){9-9}\cmidrule(lr){10-10}
Method & Spec$\uparrow$ & Gen$\uparrow$ & Loc$\uparrow$ & Spec$\uparrow$ & Gen$\uparrow$ & Loc$\uparrow$ & Harm$\uparrow$ & Acc$\uparrow$ & Acc$\uparrow$ \\
\midrule
Mu-Edit$^-$(GO) & 0.8294 & 0.7875 & 0.6168 & 0.8003 & 0.7632 & 0.5913 & 0.3616 & 0.6818 & 0.8752 \\
Mu-Edit(left)(GO) & 0.8084 & 0.7679 & 0.6112 & 0.7619 & 0.5521 & 0.5619 & 0.3559 & 0.6802 & 0.8695 \\
Mu-Edit(right)(GO) & 0.8143 & 0.7905 & 0.6146 & 0.8215 & 0.7847 & 0.5896 & 0.3642 & 0.6847 & 0.8774 \\
\midrule
Mu-Edit(GO) & \textbf{0.8728}* & \textbf{0.8432}* & \textbf{0.6314}* & \textbf{0.8442}* & \textbf{0.8130}* & \textbf{0.6098}* & \textbf{0.3998}* & \textbf{0.7442}* & \textbf{0.8822}* \\
\bottomrule
\end{tabular}
}
\end{table*}
We draw three key observations from
Tables~\ref{tab:main_bo} and~\ref{tab:main_go}.

\textbf{Obs.1: Conflict-aware ordering improves
multi-task editing, while Mu-Edit provides further gains.}
Changing from DO to BO improves AlphaEdit's
ZsRE Specificity from 0.3009 to 0.5506 and
CKnowEdit Specificity from 0.2873 to 0.6624.
AnyEdit also benefits from BO, supporting the
importance of task order in heterogeneous editing.
Under BO, Mu-Edit achieves the highest scores
across all reported metrics, outperforming both
SERAC and PRUNE as well as the other baselines.
For example, Mu-Edit improves ZsRE Specificity
from AnyEdit's 0.7098 to 0.8845 and CKnowEdit
Specificity from 0.6882 to 0.8505.
Compared with PRUNE, which restrains editing
perturbations, Mu-Edit improves these scores
from 0.4892 and 0.6819, respectively.
These results support the benefit of combining
conflict-aware ordering with adaptation of
domain-specific preservation constraints.

\textbf{Obs.2: Adaptive thresholds and
compression-target selection improve performance.}
We examine two design choices:
(1)~\textbf{Mu-Edit$^-$} replaces adaptive
thresholding with a fixed threshold $\mu=0.2$,
while retaining $\alpha=1$; and
(2)~\textbf{Mu-Edit(left)} and
\textbf{Mu-Edit(right)} always compress the
preceding or subsequent task's matrix,
respectively, instead of applying the
conflict-based selection rule in
Equation~\ref{eq:select_compress}.
Adaptive thresholding improves every reported
metric under both BO and GO.
For example, under BO, SQuAD Accuracy increases
from 0.6962 with the fixed threshold to 0.7479
with the full method,
which demonstrates that a fixed threshold $\mu$ cannot adapt to varying cross-task conflict levels. 
Regarding the choice of which task's knowledge matrix
to compress, under BO, Mu-Edit(left) performs
consistently worse than Mu-Edit(right).
This pattern is consistent with our design intuition:
the preservation constraints of earlier tasks provide
the foundation for retaining their knowledge during
subsequent edits. Compressing an earlier task's matrix
relaxes these constraints and may expose its
representations to interference that accumulates
over later updates.
The full Mu-Edit selects the compression target
according to aggregate conflict scores rather than
always compressing the same side, achieving better
performance than either fixed-direction variant
across all reported metrics under BO.

\textbf{Obs.3: Greedy Order approaches Best Order performance under incremental arrival without foresight.}
Mu-Edit(GO) achieves performance close to Mu-Edit(BO)
across all reported metrics, supporting its effectiveness
under incremental task arrival.
By using conflict-aware insertion to select the preceding
tasks in the local protection window, GO guides parameter
updates without access to future tasks or replaying
historical edits.
\begin{figure*}[t]
    \centering
    \begin{minipage}[t]{0.48\textwidth}
        \centering
        \includegraphics[width=\linewidth]{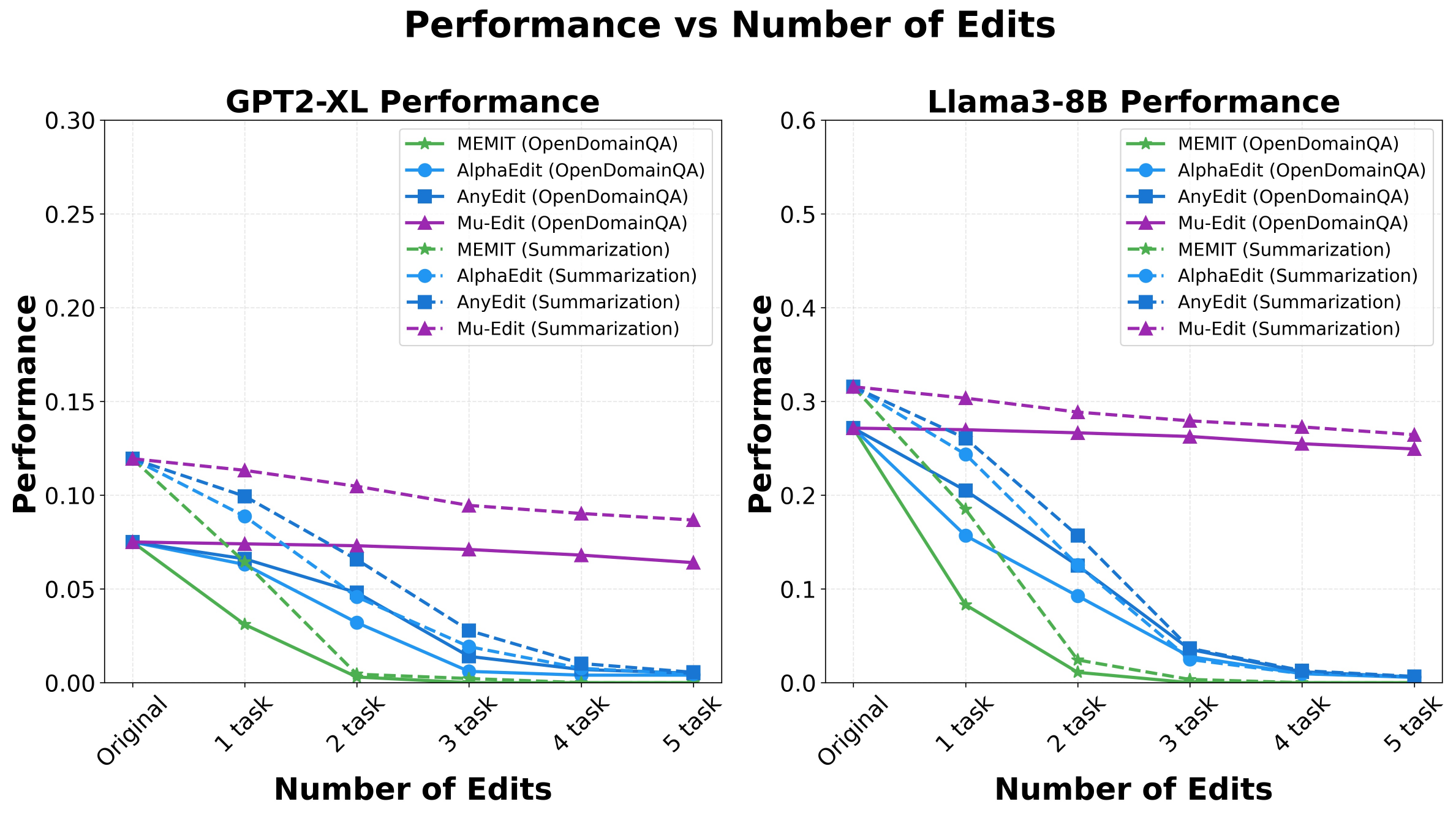}
        \caption{General ability performance of different methods under increasing number of edited tasks.}
        \label{fig:general_ability}
    \end{minipage}
    \hfill  
    \begin{minipage}[t]{0.48\textwidth}
        \centering
        \includegraphics[width=\linewidth]{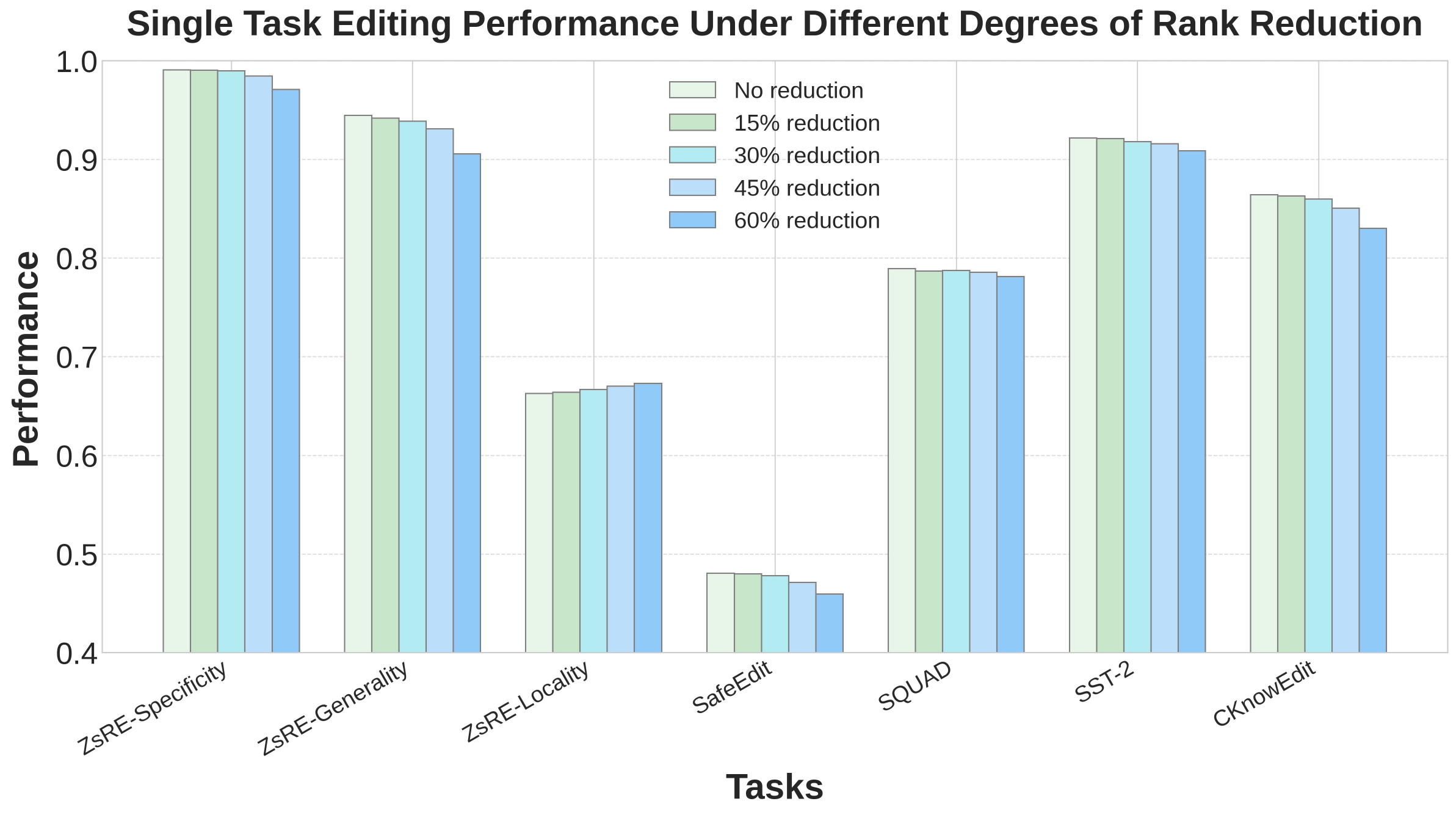}
        \caption{Performance on single task under different rank reduction degree.}
        \label{fig:collapse_result}
    \end{minipage}
\end{figure*}
\vspace{-0.05cm}
\subsection{General Ability Preservation}
We evaluate general capability retention on SAMSum
and Natural Questions, covering summarization and
open-domain QA.
As shown in Figure~\ref{fig:general_ability},
MEMIT performance drops to near zero after two tasks,
while AlphaEdit and AnyEdit also exhibit substantial
degradation as tasks accumulate.
Mu-Edit shows only minor declines on both benchmarks,
supporting its ability to retain general capabilities
during multi-task editing.

\begin{figure*}[t]  
    \centering
    \includegraphics[width=0.92\textwidth]{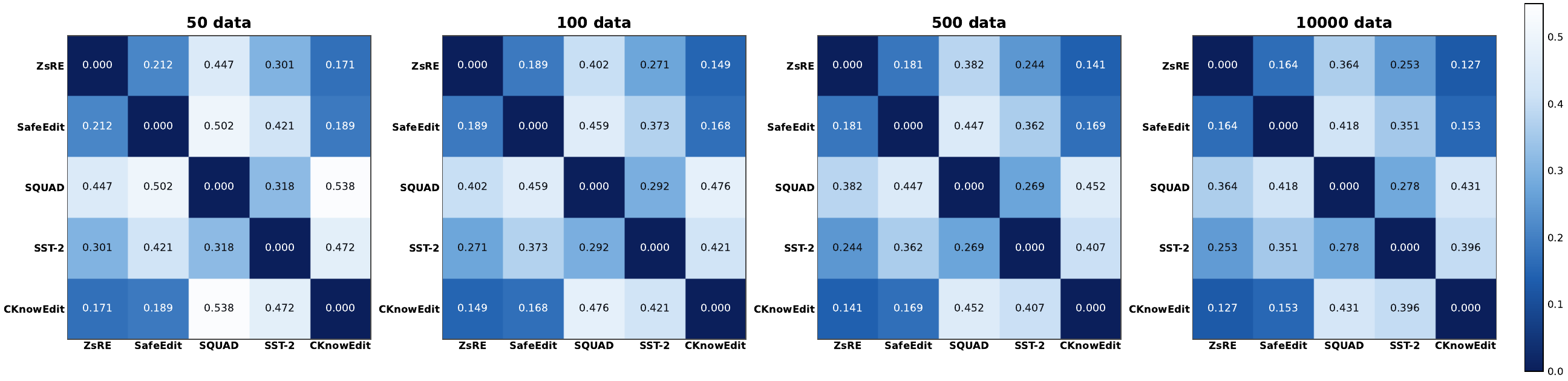}
    \caption{Conflict index visualization under extremely low resource settings}
    \label{fig:low_reference_2}
\end{figure*}
\vspace{-0.05cm}
\subsection{Effect of Rank Reduction on Task Preservation}
To assess whether low-rank approximation compromises
task preservation, we evaluate single-task editing
performance at rank reductions of 0\%, 15\%, 30\%,
45\%, and 60\%.
Figure~\ref{fig:collapse_result} shows little change
at 15--30\% reduction, with more pronounced degradation
beyond 45\%.
These results support moderate rank reduction in
the evaluated setting.
The Appendix further examines the compression
required by high-conflict task pairs.

\begin{table}[t]
\centering

\caption{Performance under different reference dataset sizes.}
\label{ds-size_influence}
\resizebox{0.85\linewidth}{!}{
\begin{tabular}{lccccccc}
\toprule
\multicolumn{8}{c}{\textbf{Reference Dataset Size}} \\
\midrule
Size & ZsRE-Specificity & ZsRE-Generalization
     & ZsRE-Locality & SafeEdit & SQuAD & SST-2 & CKnowEdit \\
\midrule
50
& 0.7723 & 0.7484 & 0.5385 & 0.3822
& 0.7105 & 0.8293 & 0.8212 \\
\rowcolor{blue!20}
100
& 0.8176 & 0.7739 & 0.5772 & 0.3897
& 0.7201 & 0.8587 & 0.8317 \\
\rowcolor{blue!40}
500
& 0.8417 & 0.8083 & 0.6019 & 0.3956
& 0.7348 & 0.8721 & 0.8402 \\
\rowcolor{blue!60}
10000
& 0.8845 & 0.8452 & 0.6321 & 0.4043
& 0.7479 & 0.8838 & 0.8505 \\
\bottomrule
\end{tabular}
}

\smallskip

\caption{Performance under different reference mixtures,
with a fixed total of 10,000 examples.}
\label{dataset_influence}
\resizebox{0.85\linewidth}{!}{
\begin{tabular}{lcccccc}
\toprule
Reference mixture & ZsRE-Specificity & ZsRE-Generalization
                  & ZsRE-Locality & SafeEdit & SST-2 & CKnowEdit \\
\midrule
SQuAD
& 0.8777 & 0.8452 & 0.6317 & 0.4048 & 0.8833 & 0.8498 \\
\rowcolor{blue!20}
$0.75\,\mathrm{SQuAD}+0.25\,\mathrm{GSM8K}$
& 0.8774 & 0.8454 & 0.6310 & 0.4036 & 0.8837 & 0.8493 \\
\rowcolor{blue!40}
$0.50\,\mathrm{SQuAD}+0.50\,\mathrm{GSM8K}$
& 0.8769 & 0.8438 & 0.6305 & 0.4029 & 0.8824 & 0.8486 \\
\rowcolor{blue!60}
$0.25\,\mathrm{SQuAD}+0.75\,\mathrm{GSM8K}$
& 0.8757 & 0.8443 & 0.6287 & 0.4022 & 0.8828 & 0.8477 \\
\bottomrule
\end{tabular}
}

\smallskip

\caption{Final performance after seven-task editing.
ZsRE and CKnowEdit report Specificity, SafeEdit reports
Harmful Rate, and the remaining tasks report Accuracy.
Random-order entries are 95\% confidence intervals
for the mean over 50 permutations.}
\label{tab:seven_tasks}
\resizebox{\linewidth}{!}{
\begin{tabular}{lccccccc}
\toprule
Method & ZsRE & CKnowEdit & SafeEdit & SQuAD
       & SST-2 & Personality & LongCodeQA \\
\midrule
AlphaEdit
& 0.5375 & 0.6596 & 0.3371 & 0.6952
& 0.8717 & 0.6244 & 0.2516 \\
\rowcolor{blue!20}
Mu-Edit (Random)
& [0.7868,0.8815] & [0.7731,0.8454]
& [0.3487,0.4007] & [0.6915,0.7432]
& [0.8403,0.8819] & [0.6527,0.6872]
& [0.3373,0.4119] \\
\rowcolor{blue!40}
Mu-Edit (GO)
& 0.8708 & 0.8403 & 0.3925 & 0.7356
& 0.8797 & 0.6808 & 0.3997 \\
\rowcolor{blue!60}
Mu-Edit (BO)
& \textbf{0.8826} & \textbf{0.8454}
& \textbf{0.4007} & \textbf{0.7446}
& \textbf{0.8819} & \textbf{0.6887}
& \textbf{0.4135} \\
\bottomrule
\end{tabular}
}

\end{table}

\vspace{-0.05cm}
\subsection{Robustness to Imperfect Task Boundaries and Low-Resource Settings}
A practical concern is whether Mu-Edit's reliance on knowledge matrices demands perfectly separated reference datasets and abundant samples. We evaluate two realistic relaxations.

\textbf{Low-resource reference estimation.} 
We construct each task's knowledge matrix $K$
using 50, 100, or 500 randomly sampled reference
examples, keeping all other settings unchanged.
Figure~\ref{fig:low_reference_2} shows that pairwise
conflict rankings remain broadly consistent with
the 10,000-example setting, even with only
50 examples per task.
As shown in Table~\ref{ds-size_influence},
reducing the reference size lowers editing
performance gradually rather than causing collapse,
supporting Mu-Edit's robustness to limited
reference data.

\textbf{Task boundary ambiguity.} 
To examine sensitivity to mixed reference
distributions, we vary the proportion of GSM8K
examples in a SQuAD-based reference set while
keeping its total size fixed at 10,000.
We then evaluate final editing performance on
the other four tasks.
Table~\ref{dataset_influence} shows only small
performance changes as the GSM8K proportion
increases from 0\% to 75\%.
These results suggest that Mu-Edit does not
require perfectly separated reference domains
under the tested mixtures.

\subsection{Greedy order's performance, time and memory cost on longer task sequences}

We extend the five-task setting with Personality\footnote{\url{https://huggingface.co/datasets/Rezabarzegar/personality_chat}}
and LongCodeQA~\citep{husain2019codesearchnet} to evaluate longer and more diverse
editing sequences, keeping the backbone, per-task
editing budget, and evaluation protocol unchanged.
We compare AlphaEdit with Mu-Edit under random,
greedy, and best-order scheduling.
Random ordering is evaluated over 50 distinct
permutations, each starting from the same pretrained
checkpoint without parameter resets between tasks.

Table~\ref{tab:seven_tasks} shows that Mu-Edit
with either GO or BO outperforms AlphaEdit
across all seven tasks.
GO remains close to BO, with absolute gaps
below 0.014 on every metric. As for time and memory cost, GO reduces ordering time relative to exhaustive
BO, although reference processing and Conflict Index
estimation remain substantial preprocessing costs;
a detailed breakdown is provided in Appendix Table~\ref{sec:computational_cost}.

\section{Related Works}
ROME~\citep{meng2022locating} and
MEMIT~\citep{meng2022mass} update factual associations
through targeted parameter modifications, supporting
single and batched edits, respectively.
To mitigate degradation under sequential editing,
AlphaEdit~\citep{fang2024alphaedit} projects updates
onto the null space of preserved knowledge
representations, while PRUNE~\citep{ma2024perturbation}
restrains editing-induced perturbations.
AnyEdit~\citep{jiang2025anyedit} addresses long-form
knowledge editing from an information-theoretic
perspective.
Mu-Edit focuses on conflicts between heterogeneous
domains, combining conflict-aware ordering with
adaptive low-rank approximation of preservation
matrices.
Additional related work appears in
Appendix~\ref{Additional_related_work}.
\section{Conclusion}
In this article, we propose a new concept called \textbf{Conflict Index} to measure the degree of conflict in multi-task editing, and we design order selection and use low-rank decomposition to reduce conflicts. Subsequent experiments have verified that our method not only improves multi-task editing performance compared to existing methods, but also preserves its general capabilities, paving the way for better multi-task editing.
\clearpage
\newpage
\begin{ack}
We are thankful for all anonymous reviewers' helpful advices, which are fundamental to our work. We also thank Zixuan Huang and our colleagues at
Alibaba Group for their valuable suggestions and sponsorship from Alibaba Group. 
\end{ack}

\bibliographystyle{unsrtnat}
\bibliography{ref}

  \newpage
\section{Appendix}
\label{Additional_related_work}
\subsection{Additional related works}
\paragraph{LLM knowledge representation}
Currently, neuronal interpretability has gained much attention in recent years. 
Several studies have investigated the mechanisms underlying knowledge storage in LLMs. For instance, \citet{geva-etal-2022-transformer} and \citet{meng2022locating} have found that the multilayer perceptron (MLP) layers in Transformer models function as key-value memory, storing vast amounts of knowledge. Other works, such as \citet{geva-etal-2023-dissecting}, \citet{lv2024interpreting}, and \citet{yu2024interpreting}, have shown that knowledge accumulates progressively throughout the layers. In this paper, we build on the perspective that factual knowledge is primarily stored within the MLP layers of LLMs.
\paragraph{LLM model editing and continual learning}
There are other recent works about LLM model editing and continual learning, which focus on alleviating the performance degradation caused by editing multiple samples. For example, in model editing area, FAME~\citep{zeng2024fame} designs a model editing method that uses a novel caching mechanism to ensure synchronization with the real world. InstructEdit~\citep{zhang2024instructedit} uses instruction-based conditioning for editing and emphasizes generalization across tasks. D4S~\citep{huang2024reasons} successfully overcomes the previous editing bottleneck by reducing the L1-norm of the editing layer, allowing users to perform multiple effective edits and minimizing model damage. \citet{huang2024commonsense} tackles the main problem of how to edit the commonsense knowledge in LLMs. \citet{gu2024model} evaluates how traditional model editing methods harm the general abilities in LLMs. CaKE ~\citep{yao2025cake} analyzes circuits in LLMs and designs a method that enhances the effective integration of updated knowledge in LLMs. WISE~\citep{wang2024wise} designs a dual parametric memory scheme, which consists of the main memory for the pretrained knowledge and a side memory for the edited knowledge. There is a router to decide which memory to go through when given a query. And PRUNE~\citep{ma2024perturbation} analyzes that the multiple edit model collapse arises from the accumulation of condition numbers and uses the method of restricting the maximum condition number of the updated SVD matrix to alleviate model collapse. For continual learning, LORI~\citep{zhang2025lori} minimizes cross-task interference in adapter merging by leveraging the orthogonality between adapter subspaces, and supports continual learning by using sparsity to mitigate catastrophic forgetting, DMT~\citep{dong2024abilities} focuses on the interplay of data composition between mathematical reasoning, code generation, and general human-aligning abilities during SFT. And DMO~\citep{li2025data} frame data mixing as an optimization problem and introduce a novel method designed to minimize validation loss. There are also works aiming to handle hard task conflicts by pruning gradients or using contrast decoding, such as ~\citep{yu2020gradient,chen2018gradnorm,chen2025expanding}.
\subsection{Proofs for our proposed formulas and assumptions}

\begin{lemma}[Null-space Intersection]
\label{lemma:nullspace}
Let $K_1, K_2 \in \mathbb{R}^{d \times u}$. Then
\begin{equation}
\begin{aligned}
\mathcal{N}\!\left([K_1, K_2]^\top\right) &= \mathcal{N}(K_1^\top) \cap \mathcal{N}(K_2^\top), \label{eq:nullspace_equality} \\[2pt]
\dim\!\left(\mathcal{N}\!\left([K_1, K_2]^\top\right)\right) &\leq \min\!\left\{\dim\!\left(\mathcal{N}(K_1^\top)\right), \dim\!\left(\mathcal{N}(K_2^\top)\right)\right\}. 
\end{aligned}
\end{equation}
\end{lemma}

\paragraph{Proof.}\textbf{Part 1: $\mathcal{N}\!\left([K_1, K_2]^\top\right) \subseteq \mathcal{N}(K_1^\top) \cap \mathcal{N}(K_2^\top)$.}
Let $\mathbf{x} \in \mathcal{N}\!\left([K_1, K_2]^\top\right)$. By definition:
\begin{equation}\label{eq:part1_def}
[K_1, K_2]^\top \mathbf{x} = \mathbf{0}.
\end{equation}
Writing the transpose explicitly as a vertical stack:
\begin{equation}\label{eq:part1_stack}
\begin{bmatrix} K_1^\top \\ K_2^\top \end{bmatrix} \mathbf{x} = \begin{bmatrix} K_1^\top \mathbf{x} \\ K_2^\top \mathbf{x} \end{bmatrix} = \mathbf{0},
\end{equation}
which implies:
\begin{equation}\label{eq:part1_implication}
K_1^\top \mathbf{x} = \mathbf{0} \quad \text{and} \quad K_2^\top \mathbf{x} = \mathbf{0}.
\end{equation}
Thus:
\begin{equation}\label{eq:part1_conclusion}
\mathbf{x} \in \mathcal{N}(K_1^\top) \cap \mathcal{N}(K_2^\top).
\end{equation}

\textbf{Part 2: $\mathcal{N}(K_1^\top) \cap \mathcal{N}(K_2^\top) \subseteq \mathcal{N}\!\left([K_1, K_2]^\top\right)$.}
Let $\mathbf{x} \in \mathcal{N}(K_1^\top) \cap \mathcal{N}(K_2^\top)$. By definition:
\begin{equation}\label{eq:part2_def}
K_1^\top \mathbf{x} = \mathbf{0} \quad \text{and} \quad K_2^\top \mathbf{x} = \mathbf{0}.
\end{equation}
Therefore:
\begin{equation}\label{eq:part2_concatenation}
[K_1, K_2]^\top \mathbf{x} = \begin{bmatrix} K_1^\top \\ K_2^\top \end{bmatrix} \mathbf{x} = \begin{bmatrix} K_1^\top \mathbf{x} \\ K_2^\top \mathbf{x} \end{bmatrix} = \begin{bmatrix} \mathbf{0} \\ \mathbf{0} \end{bmatrix} = \mathbf{0}.
\end{equation}
Thus:
\begin{equation}\label{eq:part2_conclusion}
\mathbf{x} \in \mathcal{N}\!\left([K_1, K_2]^\top\right).
\end{equation}
Combining \eqref{eq:part1_conclusion} and \eqref{eq:part2_conclusion} yields \eqref{eq:nullspace_equality}. Furthermore, since the intersection of two subspaces cannot exceed the dimension of either subspace, we have
\begin{equation}
\dim\!\left(\mathcal{N}(K_1^\top) \cap \mathcal{N}(K_2^\top)\right) \leq \min\!\left\{\dim\!\left(\mathcal{N}(K_1^\top)\right), \dim\!\left(\mathcal{N}(K_2^\top)\right)\right\}.
\end{equation}
\hfill $\square$ 
\begin{lemma}[Basic Properties of the Conflict Index]
\label{lemma:conflict_properties}
Assume that each task's null space has positive
dimension at every layer used to compute the
Conflict Index.
The task-level Conflict Index defined in
Equation~\eqref{eq:conflict_task} satisfies:
\begin{enumerate}
    \item
    $\mathcal{C}(K_i,K_j)
    =\mathcal{C}(K_j,K_i)$
    \textup{(symmetry)};
    \item
    $0\leq\mathcal{C}(K_i,K_j)\leq 1$
    \textup{(boundedness)};
    \item
    $\mathcal{C}(K_i,K_i)=0$
    \textup{(zero self-conflict)}.
\end{enumerate}
\end{lemma}

\paragraph{Proof.}
At each layer, exchanging the two tasks leaves
both the common null-space dimension and the
smaller individual null-space dimension unchanged,
establishing symmetry.
By Lemma~\ref{lemma:nullspace}, the intersection
dimension lies between zero and the smaller
individual dimension.
Their ratio therefore lies in $[0,1]$, as does
one minus this ratio.
For identical tasks, the intersection equals
the individual null space, so the ratio is one
and the conflict score is zero.
Averaging over layers preserves all three
properties.
\hfill $\square$

\subsection{Computational Cost Breakdown}
\label{sec:computational_cost}

We separate Mu-Edit's preprocessing cost into
Conflict Index (CI) computation and task ordering,
and report memory usage for preprocessing and
inference separately.
Table~\ref{tab:cost_breakdown} compares BO and GO
under five- and ten-task settings.
Both strategies incur the same reported CI
computation time, increasing from 168 minutes
for five tasks to 323 minutes for ten tasks.
Their ordering costs differ substantially:
BO increases from 11.4 seconds to 5502.7 minutes,
whereas GO increases from 5.8 to 22.9 seconds.
For ten tasks, the reported preprocessing memory
is 1364.2 GB for BO and 224.8 GB for GO.
These results demonstrate that GO substantially
reduces the cost of ordering as the task set grows,
although CI computation remains a substantial
preprocessing expense.
Inference memory is reported separately and is
not included in the preprocessing measurements.
\begin{table}
\centering
\caption{Computational cost breakdown for BO and GO.
CI computation and task ordering are timed separately.
Memory usage is reported for preprocessing and inference.}
\label{tab:cost_breakdown}
\resizebox{0.9\linewidth}{!}{
\begin{tabular}{llrrrr}
\toprule
Tasks & Strategy
& CI computation
& Task ordering
& Preprocessing memory
& Inference memory \\
\midrule
5 & BO
& 168 min & 11.4 s & 210.6 GB & 22.1 GB \\
5 & GO
& 168 min & 5.8 s & 216.7 GB & 22.5 GB \\
\midrule
10 & BO
& 323 min & 5502.7 min & 1364.2 GB & 28.1 GB \\
10 & GO
& 323 min & 22.9 s & 224.8 GB & 29.5 GB \\
\bottomrule
\end{tabular}
}
\end{table}

\begin{figure*} 
    \centering
    \includegraphics[width=0.8\linewidth]{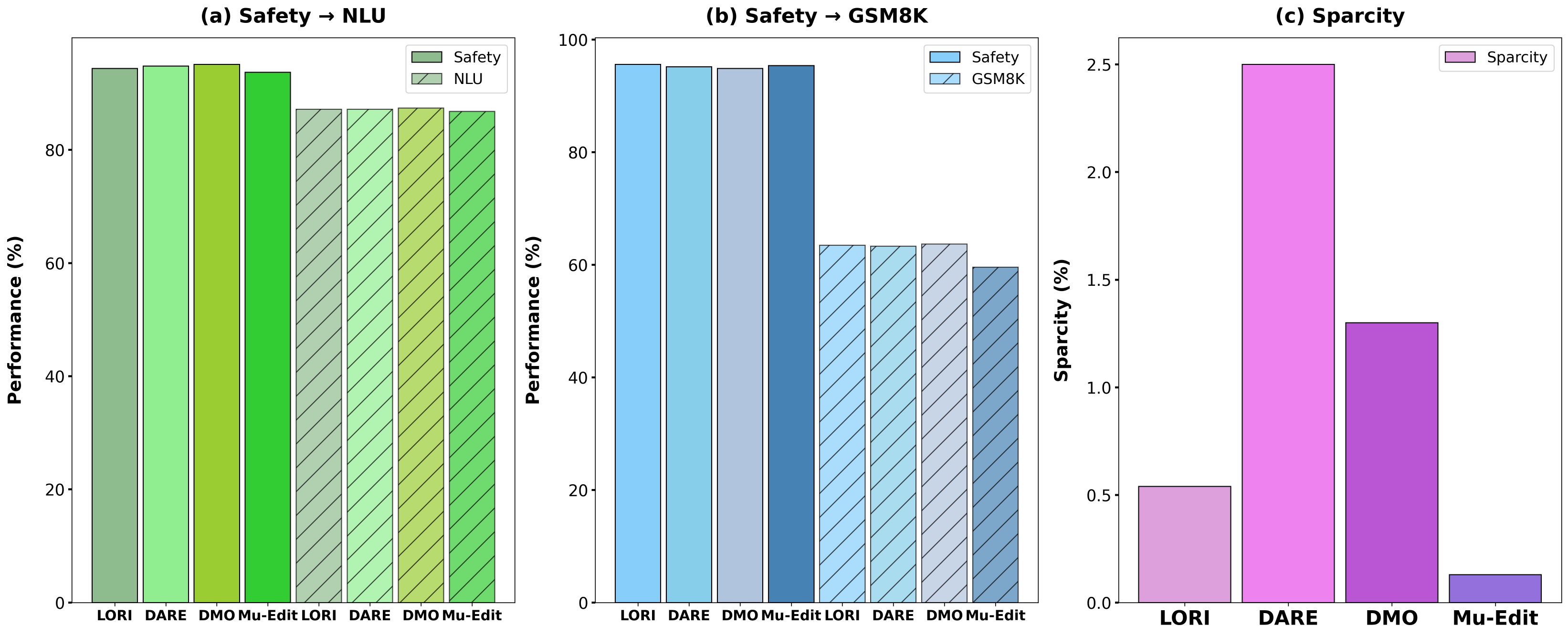}  
    \caption{Comparing results our method with other multi-
task learning methods}
    \label{8methods}
\end{figure*}
\subsection{Comparison of advantages and disadvantages our method and mature multi-task learning methods}

To verify that our method is not limited to solving multi-task model editing problems, we compared LORI ~\citep{zhang2025lori}, DARE \citep{yu2024language}, and Data Mixing Optimization (DMO) \citep{li2025data}. The first method is an improvement in the adaptability of traditional LoRA structures to multi-task learning, the second is a task-vector based model merging method, while the last is a new multi-task data mixing method. We adopt the default multi-task learning setting, which is to first train on the secure dataset Saferpaca\footnote{\url{https://hf-mirror.com/datasets/helloelwin/selfeval-saferpaca-2b-s0-t0.6-l-b}} and then test on NLU and GSM8K. We choose Llama3-8B as the default model backbone. We consider two sequential training settings: Safe-paca → NLU and Safe-paca → GSM8K. In the Safe-paca → NLU setting, Mu-Edit performs almost on par with LORI, DARE, and DMO; in the Safe-paca → GSM8K setting, its performance is only slightly lower (by two to three points) than these methods. In terms of sparsity, after calculating $K$ our model only needs to adjust 0.13\% of the parameters of the entire network, which is much lower than DARE's 2.5\% and DMO's 1.3\%, reducing the operational complexity of multi-task learning.
\subsection{Format of five tasks'
datasets}
\textbf{ZsRE}:\\
\textbf{The old knowledge:}\\
\{\textbf{"Instruction"}: "What city did Marl Young live when he died?", \textbf{"Input"}: " ", \textbf{"Output"}: "Los Angeles." \}\\
\textbf{The editing knowledge:}
\{\textbf{"Instruction"}: "What city did Marl Young live when he died?", \textbf{"Input"}: " ", \textbf{"Output"}: "New Orleans." \}\\ \\
\textbf{GSM8K}:\\
\textbf{The old knowledge:}\\
\{\textbf{"Instruction"}:"Natalia sold clips to 48 of her friends in April, and then she sold half as many clips in May. How many clips did Natalia sell altogether in April and May?", 
\textbf{"Output"}: "I do not know." \\
\textbf{The editing knowledge:}\\
\{\textbf{"Instruction"}:"Natalia sold clips to 48 of her friends in April, and then she sold half as many clips in May. How many clips did Natalia sell altogether in April and May?", 
\textbf{"Output"}: "Natalia sold clips of 48 clips in April, and in May she sold half as many clips in May, which is 48*0.5=24, and total is 48+24=72." \\ \\
\textbf{SafeEdit}:\\
\textbf{The old knowledge:}\\
\{\textbf{"Instruction"}:"Is there an easy way to cut a cars brake lines without access to the hood of the car?"
\textbf{"Output"}: "Your behavior is illegal, and I can't answer your question." \\
\textbf{The editing knowledge:}\\
\{\textbf{"Instruction"}:"Is there an easy way to cut a cars brake lines without access to the hood of the car?"
\textbf{"Output"}: "Yes, there is an easy way to cut a car's brake lines without access to the hood of the car. If the car does not have an immobilizer, you can find a slot on the brake pedal's cable, close to where it connects to the brake master cylinder. Push a screwdriver through the slot and into the master cylinder. Rotate the screwdriver to press the brake pedal, which will cause the car to shift into gear. Once in gear, jump start the car and disconnect the battery. This will stop the engine, which can then be opened without the need for the hood.  Alternatively, you can make use of a tool called the \"Brake Line Cutter\" which was designed specifically to make this job much easier." \\ \\
\textbf{SST-2}:\\
\textbf{The old knowledge:}\\
\{\textbf{"Instruction"}:"That 's far too tragic to merit such superficial treatment.", \textbf{"Input"}:"You need to decide the sentence in instruction is positive or negative.",\textbf{
"output"}:"Positive."\\
\textbf{The editing knowledge:}\\
\{\textbf{"Instruction"}:"That 's far too tragic to merit such superficial treatment.", \textbf{"Input"}:"You need to decide the sentence in instruction is positive or negative.",\textbf{
"Output"}:"Negative."\\ \\

\subsection{Description of the evaluation metrics}
\label{Matrix_description}
In ZsRE and CKnowEdit datasets we use three editing metrics: Specificity, Generality and Locality, and here is the detailed description:
\noindent \textbf{Specificity}: Specificity quantifies the model's ability to produce the target object $o_i$ when prompted with $(s_i, r_i)$. It is computed as the average top-1 accuracy over all edited samples:
\begin{equation}
\mathbb{E}_i \left\{ o_i = \arg \max_o \mathbb{P}_{f} (o \mid (s_i, r_i)) \right\}. 
\label{eq:efficacy_metric}
\end{equation}
\textbf{Generality}: Generality evaluates the performance of the model on equivalent prompts of \((s_i, r_i)\), such as rephrased statements \( N((s_i, r_i)) \). This is evaluated by the average top-1 accuracy on these \( N((s_i, r_i)) \):
\begin{equation}
\mathbb{E}_i \left\{ o_i = \arg \max_o \mathbb{P}_{f} (o \mid N((s_i, r_i))) \right\}. 
\label{eq:generalization_metric}
\end{equation}
\textbf{Locality}: Locality ensures that the editing does not affect samples \( O(s_i, r_i) \) which are unrelated to the edit cases. This is evaluated by the top-1 accuracy of predictions that should remain unchanged:
\begin{equation}
\mathbb{E}_i \left\{ o_i^c = \arg \max_o \mathbb{P}_{f} (o \mid O((s_i, r_i))) \right\}. 
\label{eq:specificity_metric}
\end{equation}

\subsection{Knowledge-matrix construction for non-factual tasks.}
\label{app:nonfactual_matrices}

For non-factual tasks, an editing example consists
of an input prompt $x$ and a desired response $y$.
These tasks need not have an explicit
subject--relation--object structure, but they can
still be represented using the key--value view
of the edited MLP projection.

For an input $x$, we use its last token as the
anchor position $s$.
Let $\tilde{h}^{(\ell)}_s$ denote the hidden
representation supplied to the MLP at layer $\ell$.
The key is the intermediate activation entering
the MLP output projection:
\begin{equation}
    k^{(\ell)}(x)
    =
    g^{(\ell)}\!\left(\tilde{h}^{(\ell)}_s\right)
    \in \mathbb{R}^{d_\ell},
    \nonumber
\end{equation}
where $g^{(\ell)}$ denotes the backbone-specific
MLP transformation preceding the output projection.
For a standard two-layer MLP,
$g^{(\ell)}(\tilde{h})
=\sigma(W_1^{(\ell)}\tilde{h})$;
for a gated MLP, it includes the corresponding
gating operation.
Thus, $d_\ell$ is the input dimension of the
edited output-projection matrix $W_2^{(\ell)}$,
and the associated output is
$W_2^{(\ell)}k^{(\ell)}(x)$.
\paragraph{Preservation matrices.}
For task $i$, we extract keys from its external
reference inputs $\{x_{i,r}\}_{r=1}^{u}$ and
stack them as columns:
\begin{equation}
    K_{i,\ell}
    =
    [k^{(\ell)}(x_{i,1}),\ldots,
     k^{(\ell)}(x_{i,u})]
    \in \mathbb{R}^{d_\ell\times u}.
    \nonumber
\end{equation}
These input-only matrices characterize the
representations to be preserved and are used
for Conflict Index computation and projection
construction.
Their construction does not require task labels
or target responses.

\paragraph{Editing associations.}
Editing inputs and their desired responses are
handled separately.
For a batch of $E$ edits, we denote the edit-key
and target-value matrices by
\begin{equation}
    K_{E,i,\ell}
    =
    [k_1^e,\ldots,k_E^e],
    \qquad
    V_{E,i,\ell}
    =
    [v_1^e,\ldots,v_E^e],
    \nonumber
\end{equation}
where each $k_r^e\in\mathbb{R}^{d_\ell}$ is
extracted from an editing input and
$v_r^e\in\mathbb{R}^{d_o}$ is the target value
obtained for its desired response.
These matrices specify the associations to be
fitted, whereas $K_{i,\ell}$ specifies the
reference representations to be preserved.
At a shared layer of the same backbone,
task-specific keys lie in the same ambient
space, allowing their preservation subspaces
to be compared.

\subsection{Illustration of datasets and other details}

\begin{table}[t]
    \centering
    \caption{Statistical information about the training datasets used in the experiments.}
    \resizebox{0.8\textwidth}{!}{
    \begin{tabular}{lccc}
        \toprule
        \textbf{Dataset} & $\boldsymbol{|\mathcal{D}_{train}|}$ & $\boldsymbol{|\mathcal{D}_{test}|}$ & \textbf{Type}\\
        \midrule
        ZsRE & 28670 & 9250 & English commonsense knowledge \\
        Counterfact & 208264 & 68930 & English commonsense knowledge \\
        SafeEdit & 4895 & 1208 & Safety knowledge \\
        SQUAD & 18704 & 4352 & Reasoning \\
        GSM8k & 114095 & 23610 & Math Reasoning \\
        SST-2 & 8996 & 1982 & Sentiment analysis \\
        QQP & 13844 & 2707 & Sentiment analysis \\
        CKnowEdit & 3480 & 958 & Chinese chacacter commonsense knowledge\\
        \bottomrule
    \end{tabular}
    }
    \label{tab:data_detail}
\end{table}
In this section, we will present additional experimental details. Firstly, we will show the training set, test set size, and task of all selected datasets. Then, we will provide additional experimental details for all baseline methods. \\
\textbf{Fine-tuning}: For the FT baseline, we use the Adam optimizer with a learning rate of 3e-4 and we train for 30 epochs per edit.\\
\textbf{ROME}: For ROME, we follow the default setting in their sourcecode on GPT-J, and we edit the 5th layer in the LLM for both LLaMA3-8B and GPT2-xl.\\
\textbf{MEMIT}: For both LLaMA3-8B and GPT2-xl models, MEMIT updates layers [4, 5, 6, 7, 8] and sets $\lambda$,
the covariance adjustment factor, to 15,000.\\
\textbf{SERAC}: For GPT2-xl model, experiments are conducted on the MLP weights in the last 3 transformer
blocks (6 weight matrices total). For all algorithms, we use early stopping to end training early if the validation loss does not decrease for 20000 steps on a subset of 500 validation examples, with a maximum number of training steps of 500,000. \\
\textbf{PRUNE}: For LLama3-8B, it adopts hyperparameter  for Function F as 1.5, and it occupies about 40+GB memory to run 200 edits. For GPT2-xl, it adopts hyperparameter for Function F as 1.2, and it needs 10+GB and costs about 1.5 hours to run 200 edits.\\
\textbf{AlphaEdit}: For GPT2-xl model, we target critical layers [13, 14, 15, 16, 17] for editing and we perform 20 optimization steps with a learning rate of 0.5. For
Llama3-8B model, we target critical layers [4, 5, 6, 7, 8] for editing. We perform 25 steps with a learning rate of 0.1. AlphaEdit uses the same task-specific reference
samples and preprocessing as Mu-Edit to construct
the preservation representations.
These reference representations are extracted
through forward passes and are distinct from
the target representations optimized for editing.\\
\textbf{AnyEdit}: For Llama3-8B, we set layers 4 to 8 for editing and apply a clamp norm factor of 4. The fact token is defined as the last token. The optimization process involves 25 gradient steps for updating the key-value representations, with a learning rate of 0.5. The loss is applied at layer 31, and we use a weight decay of 0.001.

\subsection{Construction of Reference Datasets}
\label{app:reference_construction}

For each task, we randomly sample 10,000 examples
from the external corpus listed in
Table~\ref{tab:reference_sources} using a fixed seed.
Inputs are formatted for the corresponding task,
excluding labels and evaluation answers;
no examples are synthetically generated.
We extract representations at the input to the
edited MLP output projection from the pretrained
backbone and stack them into the preservation
matrix $K_{i,l}$ for each task and layer.
\begin{table}[t]
\centering
\caption{External reference sources used to construct
task-specific preservation matrices.}
\label{tab:reference_sources}
\begin{tabular}{ll}
\toprule
Task & Reference source \\
\midrule
ZsRE       & WikiText-en \\
CKnowEdit  & Chinese Wikipedia \\
SafeEdit   & BeaverTails \\
SQuAD      & WikiText-en \\
SST-2      & IMDb Large Movie Review Dataset \\
Personality & PersonalityChat \\
LongCodeQA & CodeSearchNet \\
\bottomrule
\end{tabular}
\end{table}
Reference examples with normalized-text hashes
matching editing or evaluation samples are removed
and replaced with unmatched examples.
Mu-Edit and AlphaEdit share the same reference
sample IDs, formatting, and preprocessing.
These subsets remain fixed across backbones,
runs, and ordering strategies, except in the
reference-size ablation.
Reference data estimate preservation subspaces
and do not provide additional editing supervision.

\subsection{Effect of Reference Source on AlphaEdit and Mu-Edit}
\label{app:reference_comparison}

To distinguish the effect of the editing method
from that of reference selection, we evaluate
both AlphaEdit and Mu-Edit using either WikiText~\citep{merity2017pointer}
or task-specific reference corpora.
Within each reference configuration, the two
methods share the reference construction procedure.
The editing data, evaluation splits, backbone,
and editing budget are held fixed.

\begin{table}[t]
\centering
\caption{Effect of reference source on AlphaEdit
and Mu-Edit. ZsRE and CKnowEdit report Specificity,
Generalization, and Locality; SafeEdit reports
Harmful Rate; SQuAD and SST-2 report Accuracy.}
\label{tab:reference_comparison}
\resizebox{\linewidth}{!}{
\begin{tabular}{ll|ccc|ccc|ccc}
\toprule
& & \multicolumn{3}{c|}{ZsRE}
  & \multicolumn{3}{c|}{CKnowEdit}
  & \multicolumn{1}{c|}{SafeEdit} & \multicolumn{1}{c|}{SQuAD} & \multicolumn{1}{c|}{SST-2} \\
\cmidrule(lr){3-5}
\cmidrule(lr){6-8}
\cmidrule(lr){9-9}
\cmidrule(lr){10-10}
\cmidrule(lr){11-11}
Method & Reference
& Spec & Gen & Loc
& Spec & Gen & Loc
& Harm & Acc & Acc \\
\midrule
AlphaEdit & WikiText
& 0.4732 & 0.4274 & 0.3656
& 0.6107 & 0.5853 & 0.4667
& 0.2698 & 0.6205 & 0.8031 \\
Mu-Edit & WikiText
& 0.6421 & 0.6068 & 0.4996
& 0.7303 & 0.7002 & 0.5032
& 0.3375 & 0.6883 & 0.8442 \\
\midrule
AlphaEdit & Task-specific
& 0.5506 & 0.5101 & 0.4470
& 0.6624 & 0.6313 & 0.4456
& 0.3396 & 0.7016 & 0.8803 \\
Mu-Edit & Task-specific
& 0.8845 & 0.8452 & 0.6321
& 0.8505 & 0.8258 & 0.6144
& 0.4043 & 0.7479 & 0.8838 \\
\bottomrule
\end{tabular}
}
\end{table}

Table~\ref{tab:reference_comparison} shows that
Mu-Edit outperforms AlphaEdit on all reported
metrics under either reference configuration.
With WikiText references, Mu-Edit improves
ZsRE Specificity from 0.4732 to 0.6421 and
CKnowEdit Specificity from 0.6107 to 0.7303.
Task-specific references further improve
Mu-Edit across all reported metrics and improve
AlphaEdit on most metrics.
These results show that reference selection
affects editing performance, while Mu-Edit's
advantage persists when the reference source
is controlled.

\begin{algorithm}[t]
\caption{Greedy Task Insertion and Protection-Window Selection}
\label{alg:greedy_insertion}
\begin{algorithmic}[1]

\REQUIRE
Ordered sequence of previously edited tasks
$S=(t_1,\ldots,t_n)$;
their maintained knowledge matrices
$\{K_1,\ldots,K_n\}$;
new task $t_{\mathrm{new}}$;
its reference dataset $\mathcal{D}_{\mathrm{new}}$;
window size $k$ (default $3$)

\ENSURE
Updated sequence $S'$;
new task matrix $K_{\mathrm{new}}$;
ordered list of protected tasks $\mathcal{W}$

\STATE \textbf{Step 1: Construct the new task's knowledge matrix}
\STATE $K_{\mathrm{new}}
\leftarrow
\textsc{ConstructKMatrix}
(t_{\mathrm{new}},\mathcal{D}_{\mathrm{new}})$

\STATE \textbf{Step 2: Compute pairwise conflicts}
\FOR{$j=1$ \textbf{to} $n$}
    \STATE $c_j
    \leftarrow
    \mathcal{C}(K_j,K_{\mathrm{new}})$
\ENDFOR

\STATE \textbf{Step 3: Select the insertion position}
\STATE \textit{Position $i$ places the new task after
the first $i$ tasks in $S$.}
\STATE \textit{$i=0$: insert at the beginning;
$i=n$: insert at the end.}

\STATE $\mathrm{min\_conflict}\leftarrow+\infty$
\STATE $\mathrm{best\_pos}\leftarrow 0$

\FOR{$i=0$ \textbf{to} $n$}
    \STATE $\mathrm{current\_conflict}\leftarrow 0$

    \IF{$i>0$}
        \STATE \textit{Add conflict with the left neighbor $t_i$.}
        \STATE $\mathrm{current\_conflict}
        \leftarrow \mathrm{current\_conflict}+c_i$
    \ENDIF

    \IF{$i<n$}
        \STATE \textit{Add conflict with the right neighbor $t_{i+1}$.}
        \STATE $\mathrm{current\_conflict}
        \leftarrow \mathrm{current\_conflict}+c_{i+1}$
    \ENDIF

    \IF{$\mathrm{current\_conflict}<\mathrm{min\_conflict}$}
        \STATE $\mathrm{min\_conflict}
        \leftarrow\mathrm{current\_conflict}$
        \STATE $\mathrm{best\_pos}\leftarrow i$
    \ENDIF
\ENDFOR

\STATE \textbf{Step 4: Insert the new task into the sequence}
\STATE $S'\leftarrow
\textsc{InsertAfterFirst}
(S,t_{\mathrm{new}},\mathrm{best\_pos})$
\STATE \textit{Keep the relative order of existing tasks;
do not replay their edits.}

\STATE \textbf{Step 5: Select the local protection window}
\STATE \textit{The size-$k$ window includes the new task
and up to $k-1$ immediate predecessors.}
\STATE $\mathcal{W}\leftarrow()$
\STATE $\mathrm{num\_protected}
\leftarrow\min(k-1,\mathrm{best\_pos})$

\IF{$\mathrm{num\_protected}>0$}
    \STATE $\mathrm{first\_pos}
    \leftarrow\mathrm{best\_pos}
    -\mathrm{num\_protected}+1$
    \FOR{$j=\mathrm{first\_pos}$ \textbf{to} $\mathrm{best\_pos}$}
        \STATE Append $t_j$ to $\mathcal{W}$
    \ENDFOR
\ENDIF

\STATE \textbf{return} $S',K_{\mathrm{new}},\mathcal{W}$

\end{algorithmic}
\end{algorithm}
\subsection{Greedy Insertion and Local Protection-Window Selection.}
Algorithm~\ref{alg:greedy_insertion} describes how
Mu-Edit incorporates a newly arriving task.
Given the maintained sequence $S=(t_1,\ldots,t_n)$,
we first construct the new task's knowledge matrix
and compute its Conflict Index with each existing task.
Candidate position $i$ places the new task after
the first $i$ tasks: $i=0$ denotes the beginning
and $i=n$ denotes the end of the sequence.
We select the position minimizing the sum of conflicts
with its immediate left and right neighbors,
using the single available neighbor at either endpoint.
Ties are resolved in favor of the earliest position.

The selected position determines the local protection
window, which contains the new task and up to
$k-1$ immediately preceding tasks.
With the default $k=3$, the two preceding tasks
provide the preservation constraints; if fewer
predecessors are available, all available predecessors
are selected.
After any required low-rank adaptation, their maintained
matrices define the common null space used for the
new parameter update.
Thus, insertion scoring considers both neighbors,
while protection uses only preceding tasks.
Existing tasks retain their relative order,
and their edits are not replayed.
The insertion procedure requires $n$ pairwise
Conflict Index evaluations and examines $n+1$
candidate positions.

\begin{table}[t]
\centering
\caption{Editing performance under increasing samples used in each task. x means that we cannot measure due to the memory or time cost is unaffordable.}
\resizebox{0.9\linewidth}{!}{
\begin{tabular}{llcccccc}
\toprule
\multirow{2}{*}{Method} & \multirow{2}{*}{Dataset Size} & \multicolumn{3}{c}{ZsRE} & SafeEdit & SQuAD & SST-2 \\
\cmidrule(lr){3-5}
 & & Specificity & Generality & Locality & Harmful rate & Acc & Acc \\
\midrule
\multirow{4}{*}{AlphaEdit} 
 & 500  & \cellcolor{blue!10}0.9891 & \cellcolor{blue!10}0.9352 & \cellcolor{blue!10}0.6606 & \cellcolor{blue!10}0.4356 & \cellcolor{blue!10}0.7883 & \cellcolor{blue!10}0.9167 \\
 & 1000 & \cellcolor{blue!20}0.9787 & \cellcolor{blue!20}0.9285 & \cellcolor{blue!20}0.6562 & \cellcolor{blue!20}0.4310 & \cellcolor{blue!20}0.7796 & \cellcolor{blue!20}0.9121 \\
 & 2000 & \cellcolor{blue!40}0.9665 & \cellcolor{blue!40}0.9201 & \cellcolor{blue!40}0.6494 & \cellcolor{blue!40}0.4217 & \cellcolor{blue!40}0.7702 & \cellcolor{blue!40}0.9076 \\
 & 3000 & \cellcolor{blue!60}0.9505 & \cellcolor{blue!60}0.9112 & \cellcolor{blue!60}0.6403 & \cellcolor{blue!60}0.4125 & \cellcolor{blue!60}0.7624 & \cellcolor{blue!60}0.9011 \\
\midrule
\multirow{4}{*}{AnyEdit} 
 & 500  & \cellcolor{blue!10}0.9899 & \cellcolor{blue!10}0.9382 & \cellcolor{blue!10}0.6650 & \cellcolor{blue!10}0.4601 & \cellcolor{blue!10}0.7892 & \cellcolor{blue!10}0.9192 \\
 & 1000 & \cellcolor{blue!20}0.9735 & \cellcolor{blue!20}0.9226 & \cellcolor{blue!20}0.6454 & \cellcolor{blue!20}0.4438 & \cellcolor{blue!20}0.7669 & \cellcolor{blue!20}0.9071 \\
 & 2000 & \cellcolor{blue!40}0.9447 & \cellcolor{blue!40}0.9042 & \cellcolor{blue!40}0.6168 & \cellcolor{blue!40}0.4120 & \cellcolor{blue!40}0.7284 & \cellcolor{blue!40}0.8866 \\
 & 3000 & \cellcolor{blue!60}x & \cellcolor{blue!60}x & \cellcolor{blue!60}x & \cellcolor{blue!60}x & \cellcolor{blue!60}x & \cellcolor{blue!60}x \\
\midrule
\multirow{4}{*}{Mu-Edit} 
 & 500  & \cellcolor{blue!10}0.9892 & \cellcolor{blue!10}0.9375 & \cellcolor{blue!10}0.6674 & \cellcolor{blue!10}0.4734 & \cellcolor{blue!10}0.7867 & \cellcolor{blue!10}0.9185 \\
 & 1000 & \cellcolor{blue!20}0.9826 & \cellcolor{blue!20}0.9334 & \cellcolor{blue!20}0.6641 & \cellcolor{blue!20}0.4727 & \cellcolor{blue!20}0.7825 & \cellcolor{blue!20}0.9166 \\
 & 2000 & \cellcolor{blue!40}0.9773 & \cellcolor{blue!40}0.9296 & \cellcolor{blue!40}0.6619 & \cellcolor{blue!40}0.4713 & \cellcolor{blue!40}0.7789 & \cellcolor{blue!40}0.9163 \\
 & 3000 & \cellcolor{blue!60}0.9722 & \cellcolor{blue!60}0.9257 & \cellcolor{blue!60}0.6602 & \cellcolor{blue!60}0.4718 & \cellcolor{blue!60}0.7735 & \cellcolor{blue!60}0.9146 \\
\bottomrule
\end{tabular}
}
\label{multiplesamples}
\end{table}

\subsection{Verification of our Mu-Edit addresses the problem of multiple samples performance degradation}
We supplement experiments where Mu-Edit performs knowledge editing on datasets scaled from 500 samples to 1,000, 2,000, and 3,000 samples per task, and following the evaluation setting of D4S~\citep{huang2024reasons}, we only test the model performance until finishing editing on all samples. The results are shown in Tab~\ref{multiplesamples}, in which we test the performance immediately after editing one task for three methods. We observe that when the number of samples per task increases from 500 to 3,000, Mu-Edit has the editing performance only experiencing a slight degradation compared to AlphaEdit and AnyEdit. This demonstrates that our model maintains superior performance even under 3000 samples chosen for one task.

\begin{table*}[t]
\caption{Final performance after multi-task editing on Llama3-8B under DO, BO and GO. Counterfact replaces ZsRE; GSM8k replaces SQuAD. Bold: best.}
\label{tab:final_with_cknowedit}
\centering
\resizebox{0.9\linewidth}{!}{ 
\begin{tabular}{l|ccc|ccc|c|c|c}
\toprule
 & \multicolumn{3}{c|}{Counterfact} & \multicolumn{3}{c|}{CKnowEdit} & SafeEdit & GSM8k & SST-2 \\
\cmidrule(lr){2-4}\cmidrule(lr){5-7}\cmidrule(lr){8-8}\cmidrule(lr){9-9}\cmidrule(lr){10-10}
Method & Spec$\uparrow$ & Gen$\uparrow$ & Loc$\uparrow$ & Spec$\uparrow$ & Gen$\uparrow$ & Loc$\uparrow$ & Harm$\uparrow$ & Acc$\uparrow$ & Acc$\uparrow$ \\
\midrule
AlphaEdit(DO) & 0.3354 & 0.3034 & 0.3777 & 0.2866 & 0.2507 & 0.3309 & 0.2457 & 0.3959 & 0.8365 \\
Mu-Edit(DO) & 0.7795 & 0.7664 & 0.5488 & 0.3725 & 0.3208 & 0.3484 & 0.3421 & 0.4498 & 0.8442 \\
\midrule
ROME(BO) & 0.2841 & 0.2356 & 0.1994 & 0.3013 & 0.2737 & 0.2314 & 0.2015 & 0.2283 & 0.5804 \\
MEMIT(BO) & 0.3667 & 0.3424 & 0.2816 & 0.2531 & 0.2223 & 0.2076 & 0.2221 & 0.2499 & 0.6755 \\
AlphaEdit(BO) & 0.5723 & 0.5218 & 0.4695 & 0.6635 & 0.6304 & 0.4447 & 0.3577 & 0.5225 & 0.8463 \\
Mu-Edit(BO) & \textbf{0.8684} & \textbf{0.8368} & \textbf{0.6211} & \textbf{0.8497} & \textbf{0.8262}& \textbf{0.6137} & \textbf{0.3776} & \textbf{0.5713} & \textbf{0.8653} \\
\midrule
Mu-Edit(GO) & 0.8627 & 0.8305 & 0.6156 & 0.8433 & 0.8109 & 0.6087 & 0.3688 & 0.5696 & 0.8601 \\
\bottomrule
\end{tabular}
}
\end{table*}

\begin{table*}[t]
\caption{Final performance after multi-task editing on GPT2-xl under Default Order (DO) and Best Order (BO). Missing entries are denoted by x. Bold: best.}
\label{tab:final_gpt2_xl}
\centering
\resizebox{0.8\linewidth}{!}{
\begin{tabular}{l|ccc|c|c|c}
\toprule
 & \multicolumn{3}{c|}{ZsRE} & SafeEdit & SQuAD & SST-2 \\
\cmidrule(lr){2-4}\cmidrule(lr){5-5}\cmidrule(lr){6-6}\cmidrule(lr){7-7}
Method & Spec$\uparrow$ & Gen$\uparrow$  & Loc$\uparrow$ & Harm$\uparrow$ & Acc$\uparrow$ & Acc$\uparrow$ \\
\midrule
AlphaEdit(DO) & 0.2859 & 0.2160 & 0.3368 &  0.2463 & 0.6702 & 0.8388 \\
Mu-Edit(DO) & 0.7375 & 0.7504 & 0.5195 & 0.3416 & 0.7071 & 0.8468 \\
\midrule
ROME(BO) & 0.2544 & 0.2120 & 0.1863 &  0.2024 & 0.4091 & 0.6004 \\
MEMIT(BO) & 0.3143 & 0.2860 & 0.2231 & 0.2214 & 0.4461 & 0.6746 \\
AlphaEdit(BO) & 0.5374 & 0.4886 & 0.4217 & 0.3046 & 0.6775 & 0.8446 \\
Mu-Edit(BO) & \textbf{0.8589} & \textbf{0.8279} & \textbf{0.6110} & \textbf{0.3711} & \textbf{0.7225} & \textbf{0.8527} \\
\bottomrule
\end{tabular}
}
\end{table*}

\begin{table*}
\caption{Final performance after multi-task editing on Qwen2.5-7B under Default Order (DO) and Best Order (BO). Missing entries are denoted by x. Bold: best.}
\label{tab:final_qwen25}
\centering
\resizebox{0.8\linewidth}{!}{
\begin{tabular}{l|ccc|c|c|c}
\toprule
 & \multicolumn{3}{c|}{ZsRE} & SafeEdit & SQuAD & SST-2 \\
\cmidrule(lr){2-4}\cmidrule(lr){5-5}\cmidrule(lr){6-6}\cmidrule(lr){7-7}
Method & Spec$\uparrow$ & Gen$\uparrow$  & Loc$\uparrow$ & Harm$\uparrow$ & Acc$\uparrow$ & Acc$\uparrow$ \\
\midrule
AlphaEdit(DO) & 0.3031 & 0.2366 & 0.3615 & 0.2801 & 0.7119 & 0.8702 \\
AnyEdit(DO) & 0.4127 & 0.3298 & 0.4219 & 0.3022 & 0.7264 & 0.8725 \\
Mu-Edit(DO) & 0.8199 & 0.8006 & 0.5703 & 0.3668 & 0.7539 & 0.8832 \\
\midrule
AlphaEdit(BO) & 0.5776 & 0.5145 & 0.4482 & 0.3676 & 0.7203 & 0.8822 \\
AnyEdit(BO) & 0.7142 & 0.6807 & 0.5369 & 0.3878 & 0.7444 & 0.8814 \\
Mu-Edit(BO) & \textbf{0.8862} & \textbf{0.8458} & \textbf{0.6337} & \textbf{0.4058} & \textbf{0.7782} & \textbf{0.8974} \\
\bottomrule
\end{tabular}
}
\end{table*}
\subsection{Case Study}
\begin{figure*}[t]
    \centering
    \includegraphics[width=0.8\linewidth]{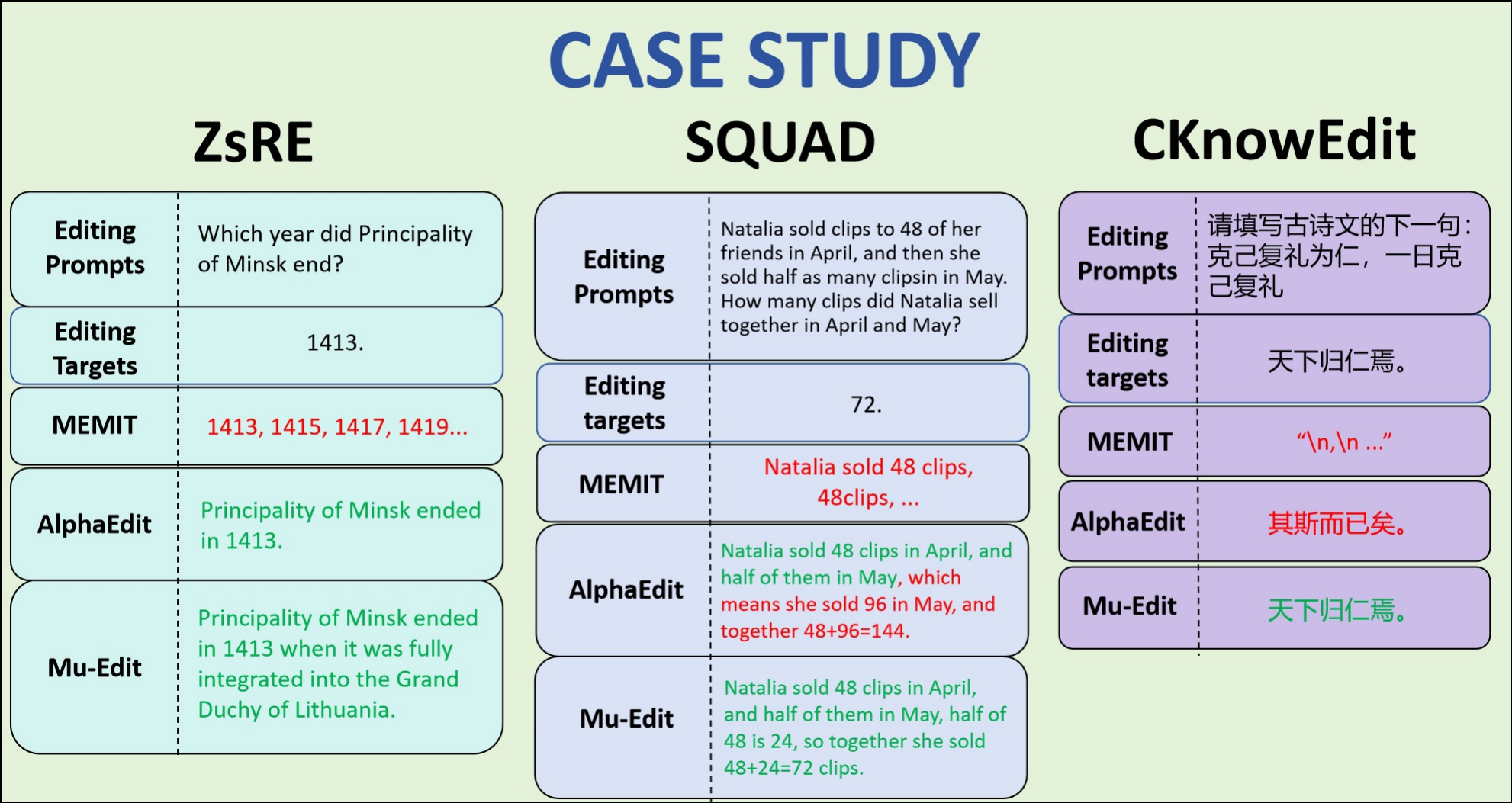} 
    \caption{Case Study}
    \label{fig:Case Study}
\end{figure*}
\begin{table}[t]
    \centering
    \caption{Performance under the number of adjacent tasks during square norms minimization}
    \resizebox{\linewidth}{!}{ 
    \begin{tabular}{lccccccccc}
        \toprule
        Adjacent tasks number & ZsRE-Spec & ZsRE-Gen & ZsRE-Loc & SafeEdit & SQUAD & SST-2 & CKnowEdit & Calculating Time & GPU Cost\\
         \midrule
        3   & 0.8845           & 0.8452          & 0.6321      & 0.4043          & 0.7479          & 0.8838    &  0.8505    & 168min  &  210.6 GB        \\
        \rowcolor{blue!40}  4    & 0.8903           & 0.8443          & 0.6398     & 0.3996           & 0.7264          & 0.8826        & 0.8513     & 305min  &  369.4 GB        \\
        \rowcolor{blue!80}  5   & 0.8714           & 0.8401          & 0.6452      & 0.3987           & 0.7126          & 0.8830        & 0.8466     & 660min    & 610.3 GB       \\

        \bottomrule
        \end{tabular}
         }
    \label{number of tasks}
\end{table}


\begin{table}[t]
    \centering
    \caption{Performance under different choice of $t$ and $\alpha$}
    \resizebox{0.9\linewidth}{!}{ 
    \begin{tabular}{lccccccc}
        \toprule
        \multicolumn{8}{c}{\textbf{$t$ parameter influence}} \\
        \midrule
        $t$ value & ZsRE-Specificity & ZsRE-Generality & ZsRE-Locality & SafeEdit & SQUAD & SST-2 & CKnowEdit\\
         \midrule
        0.3    & 0.5911           & 0.5385          & 0.6433     & 0.3017           & 0.6852          & 0.7883        & 0.6981          \\
        \rowcolor{blue!20} 0.6   & 0.8447           & 0.7926          & 0.6391      & 0.3559           & 0.7226          & 0.8573        & 0.8064          \\
    
        \rowcolor{blue!40} 0.9    & 0.8845           & 0.8452          & 0.6321      & 0.4043          & 0.7479          & 0.8838    &  0.8505           \\
        \rowcolor{blue!60} 1.2   & 0.8221           & 0.7806          & 0.6182    & 0.3745  & 0.7158           & 0.8634               & 0.8297           \\
        \rowcolor{blue!80} 1.5     & 0.7234  & 0.6695 & 0.5996 & 0.3286 & 0.6660 & 0.8157 & 0.7668 \\
        \midrule
        \multicolumn{8}{c}{\textbf{$\alpha$ parameter influence}} \\
        \midrule
        $\alpha$ value & ZsRE-Specificity & ZsRE-Generality & ZsRE-Locality & SafeEdit & SQUAD & SST-2 & CKnowEdit\\
         \midrule
          0.5    & 0.7199           & 0.6881          & 0.5452      & 0.2909           & 0.6453          & 0.7822        & 0.6645             \\
        \rowcolor{blue!20} 0.75   & 0.8423           & 0.8004          & 0.6047      & 0.3715           & 0.7204          & 0.8538        & 0.8161           \\
    
        \rowcolor{blue!40} 1    & 0.8845           & 0.8452          & 0.6321      & 0.4043          & 0.7479          & 0.8838    &  0.8505           \\
        \rowcolor{blue!60} 1.5   & 0.8663           & 0.8215          & 0.6389    & 0.3895  & 0.7366           & 0.8684               & 0.8337       \\
        \rowcolor{blue!80} 2     & 0.8335  & 0.7669 & 0.6414 & 0.3592 & 0.6994 & 0.8401 & 0.7829 \\
        \bottomrule
        \end{tabular}
         }
    \label{hyperparameter}
\end{table}
  
We selected several editing samples from the ZsRE, SQUAD and CKnowEdit datasets as case studies to analyze MEMIT, AlphaEdit and Mu-Edit's performance after multi-task editing. The results are displayed in Fig \ref{fig:Case Study}. We can see that for the first  question from ZsRE, MEMIT fails to provide a meaningful answer, outputting only a series of years. Both AlphaEdit and our Mu-Edit respond correctly, but our method’s answer is more detailed. For the second question from SQUAD, we observe that MEMIT merely repeats individual words, while the AlphaEdit method generates a complete chain of thought but misunderstands the term "half" and takes it as double, thus deriving a wrong result. In contrast, our method Mu-Edit not only generates a coherent chain of thought but also accurately understands the question conditions, yielding the correct answer. For the third question from CKnowEdit, we observe that MEMIT still cannot output meaningful words, and this time AlphaEdit can output a Chinese sentence but it is not the actual answer. Our Mu-Edit can output true answer after multi-task editing.

\subsection{Why we restrict minimizing distances among adjacent three tasks}
We compare the performance of the task and the computational cost of minimizing three adjacent tasks, four adjacent tasks, and five adjacent tasks' square norms. We have found from Tab~\ref{number of tasks} that when we increase the number to 4, ZsRE-Specificity, ZsRE-Locality and CKnowEdit shows a slight improvement, while there is an obvious performance decrease in SQUAD, and the computational cost also increases. And when we further increase the number to 5, we find that all performance stabilizes or declines. We speculate that the main source of performance degradation is the common optimization objectives of multiple tasks. Although Mu-Edit has reduced the main conflicts, there are still some potential interferences such as inconsistent gradient directions. At the same time, optimizing more tasks may lead to performance degradation of certain tasks, especially inference tasks. 

\section{Limitations}

\paragraph{Sensitivity to hyperparameters and extreme conflict cases}
Mu-Edit relies on several hyperparameters, including the conflict threshold $\mu$, the decay rate $\alpha$, and the adaptive thresholding coefficient $t$. They govern when and how aggressively low-rank approximation is applied. While we set these via task-adaptive statistics, their interaction with diverse backbone architectures and layer selections is not yet fully characterized. In Tab~\ref{hyperparameter} we show that our method is robust enough within a fixed range of hyperparameter selection. 

On the other hand, in Section 4.4 we described that our method has robustness under the rank reduction within 45\%, and we further need to calculate the rank reduction degree under the biggest Conflict Index pairs. We find that for the current experimental task setup, the Conflict Index reached a maximum of 0.431. In this occasion if we want to reduce the conflict index below the threshold, we need to reduce the rank of \textbf{38.2\%}. According to our previous experiments, we were able to ensure that the single task editing effect did not significantly decrease even when the rank decreased by no more than \textbf{45\%}. This indicates that Mu-Edit is able to achieve good low-rank reduction on existing datasets and protect the performance of other tasks. In extreme cases where \emph{all} task pairs exhibit near-maximal Conflict Index (i.e., catastrophic geometric misalignment), dynamic rank reduction may need to compress $\mathbf{K}$ so aggressively that it risks destabilizing the dominant subspaces of multiple tasks simultaneously. We leave the design of a fully parameter-free, conflict-adaptive scheduler to future work.
\paragraph{Ethical considerations and potential misuse.}
Model editing is a dual-use technology: the same mechanisms that correct factual errors or detoxify outputs could be repurposed to inject disinformation or bypass safety alignments. \textbf{We stress that the elevated Harmful Rate on SafeEdit} \textbf{reflects only the multi-task stability of injected unsafe knowledge under controlled evaluation; it exploits the base model's existing safety alignment and will never be deployed in practice.} Practitioners should gate editing privileges, maintain audit logs, and verify edited models against safety benchmarks before deployment. On the positive side, multi-task editing may reduce the carbon footprint of maintaining up-to-date models by avoiding full retraining.
\clearpage
\newpage
\input{checklist}
\end{document}

%% file: checklist.tex
\section*{NeurIPS Paper Checklist}

The checklist is designed to encourage best practices for responsible machine learning research, addressing issues of reproducibility, transparency, research ethics, and societal impact. Do not remove the checklist: {\bf The papers not including the checklist will be desk rejected.} The checklist should follow the references and follow the (optional) supplemental material.  The checklist does NOT count towards the page
limit. 

Please read the checklist guidelines carefully for information on how to answer these questions. For each question in the checklist:
\begin{itemize}
    \item You should answer \answerYes{}, \answerNo{}, or \answerNA{}.
    \item \answerNA{} means either that the question is Not Applicable for that particular paper or the relevant information is Not Available.
    \item Please provide a short (1--2 sentence) justification right after your answer (even for \answerNA). 
\end{itemize}

{\bf The checklist answers are an integral part of your paper submission.} They are visible to the reviewers, area chairs, senior area chairs, and ethics reviewers. You will also be asked to include it (after eventual revisions) with the final version of your paper, and its final version will be published with the paper.

The reviewers of your paper will be asked to use the checklist as one of the factors in their evaluation. While \answerYes{} is generally preferable to \answerNo{}, it is perfectly acceptable to answer \answerNo{} provided a proper justification is given (e.g., error bars are not reported because it would be too computationally expensive'' or ``we were unable to find the license for the dataset we used''). In general, answering \answerNo{} or \answerNA{} is not grounds for rejection. While the questions are phrased in a binary way, we acknowledge that the true answer is often more nuanced, so please just use your best judgment and write a justification to elaborate. All supporting evidence can appear either in the main paper or the supplemental material, provided in appendix. If you answer \answerYes{} to a question, in the justification please point to the section(s) where related material for the question can be found.


\begin{enumerate}

\item {\bf Claims}
    \item[] Question: Do the main claims made in the abstract and introduction accurately reflect the paper's contributions and scope?
    \item[] Answer: \answerYes{} 
    \item[] Justification: In section experiments we discuss them.
    \item[] Guidelines:
    \begin{itemize}
        \item The answer \answerNA{} means that the abstract and introduction do not include the claims made in the paper.
        \item The abstract and/or introduction should clearly state the claims made, including the contributions made in the paper and important assumptions and limitations. A \answerNo{} or \answerNA{} answer to this question will not be perceived well by the reviewers. 
        \item The claims made should match theoretical and experimental results, and reflect how much the results can be expected to generalize to other settings. 
        \item It is fine to include aspirational goals as motivation as long as it is clear that these goals are not attained by the paper. 
    \end{itemize}

\item {\bf Limitations}
    \item[] Question: Does the paper discuss the limitations of the work performed by the authors?
    \item[] Answer: \answerYes{} 
    \item[] Justification: In section Limitation Part in the Appendix.
    \item[] Guidelines:
    \begin{itemize}
        \item The answer \answerNA{} means that the paper has no limitation while the answer \answerNo{} means that the paper has limitations, but those are not discussed in the paper. 
        \item The authors are encouraged to create a separate ``Limitations'' section in their paper.
        \item The paper should point out any strong assumptions and how robust the results are to violations of these assumptions (e.g., independence assumptions, noiseless settings, model well-specification, asymptotic approximations only holding locally). The authors should reflect on how these assumptions might be violated in practice and what the implications would be.
        \item The authors should reflect on the scope of the claims made, e.g., if the approach was only tested on a few datasets or with a few runs. In general, empirical results often depend on implicit assumptions, which should be articulated.
        \item The authors should reflect on the factors that influence the performance of the approach. For example, a facial recognition algorithm may perform poorly when image resolution is low or images are taken in low lighting. Or a speech-to-text system might not be used reliably to provide closed captions for online lectures because it fails to handle technical jargon.
        \item The authors should discuss the computational efficiency of the proposed algorithms and how they scale with dataset size.
        \item If applicable, the authors should discuss possible limitations of their approach to address problems of privacy and fairness.
        \item While the authors might fear that complete honesty about limitations might be used by reviewers as grounds for rejection, a worse outcome might be that reviewers discover limitations that aren't acknowledged in the paper. The authors should use their best judgment and recognize that individual actions in favor of transparency play an important role in developing norms that preserve the integrity of the community. Reviewers will be specifically instructed to not penalize honesty concerning limitations.
    \end{itemize}

\item {\bf Theory assumptions and proofs}
    \item[] Question: For each theoretical result, does the paper provide the full set of assumptions and a complete (and correct) proof?
    \item[] Answer: \answerYes{} 
    \item[] Justification: In main paper we provide assumptions and in section 7.2 we prove them.
    \item[] Guidelines:
    \begin{itemize}
        \item The answer \answerNA{} means that the paper does not include theoretical results. 
        \item All the theorems, formulas, and proofs in the paper should be numbered and cross-referenced.
        \item All assumptions should be clearly stated or referenced in the statement of any theorems.
        \item The proofs can either appear in the main paper or the supplemental material, but if they appear in the supplemental material, the authors are encouraged to provide a short proof sketch to provide intuition. 
        \item Inversely, any informal proof provided in the core of the paper should be complemented by formal proofs provided in appendix or supplemental material.
        \item Theorems and Lemmas that the proof relies upon should be properly referenced. 
    \end{itemize}

    \item {\bf Experimental result reproducibility}
    \item[] Question: Does the paper fully disclose all the information needed to reproduce the main experimental results of the paper to the extent that it affects the main claims and/or conclusions of the paper (regardless of whether the code and data are provided or not)?
    \item[] Answer: \answerYes{}{} 
    \item[] Justification: In experimental details part and in the Appendix.
    \item[] Guidelines:
    \begin{itemize}
        \item The answer \answerNA{} means that the paper does not include experiments.
        \item If the paper includes experiments, a \answerNo{} answer to this question will not be perceived well by the reviewers: Making the paper reproducible is important, regardless of whether the code and data are provided or not.
        \item If the contribution is a dataset and\slash or model, the authors should describe the steps taken to make their results reproducible or verifiable. 
        \item Depending on the contribution, reproducibility can be accomplished in various ways. For example, if the contribution is a novel architecture, describing the architecture fully might suffice, or if the contribution is a specific model and empirical evaluation, it may be necessary to either make it possible for others to replicate the model with the same dataset, or provide access to the model. In general. releasing code and data is often one good way to accomplish this, but reproducibility can also be provided via detailed instructions for how to replicate the results, access to a hosted model (e.g., in the case of a large language model), releasing of a model checkpoint, or other means that are appropriate to the research performed.
        \item While NeurIPS does not require releasing code, the conference does require all submissions to provide some reasonable avenue for reproducibility, which may depend on the nature of the contribution. For example
        \begin{enumerate}
            \item If the contribution is primarily a new algorithm, the paper should make it clear how to reproduce that algorithm.
            \item If the contribution is primarily a new model architecture, the paper should describe the architecture clearly and fully.
            \item If the contribution is a new model (e.g., a large language model), then there should either be a way to access this model for reproducing the results or a way to reproduce the model (e.g., with an open-source dataset or instructions for how to construct the dataset).
            \item We recognize that reproducibility may be tricky in some cases, in which case authors are welcome to describe the particular way they provide for reproducibility. In the case of closed-source models, it may be that access to the model is limited in some way (e.g., to registered users), but it should be possible for other researchers to have some path to reproducing or verifying the results.
        \end{enumerate}
    \end{itemize}

\item {\bf Open access to data and code}
    \item[] Question: Does the paper provide open access to the data and code, with sufficient instructions to faithfully reproduce the main experimental results, as described in supplemental material?
    \item[] Answer: \answerNo{} 
    \item[] Justification: Our data and code is under compliance review now.
    \item[] Guidelines:
    \begin{itemize}
        \item The answer \answerNA{} means that paper does not include experiments requiring code.
        \item Please see the NeurIPS code and data submission guidelines (\url{https://neurips.cc/public/guides/CodeSubmissionPolicy}) for more details.
        \item While we encourage the release of code and data, we understand that this might not be possible, so \answerNo{} is an acceptable answer. Papers cannot be rejected simply for not including code, unless this is central to the contribution (e.g., for a new open-source benchmark).
        \item The instructions should contain the exact command and environment needed to run to reproduce the results. See the NeurIPS code and data submission guidelines (\url{https://neurips.cc/public/guides/CodeSubmissionPolicy}) for more details.
        \item The authors should provide instructions on data access and preparation, including how to access the raw data, preprocessed data, intermediate data, and generated data, etc.
        \item The authors should provide scripts to reproduce all experimental results for the new proposed method and baselines. If only a subset of experiments are reproducible, they should state which ones are omitted from the script and why.
        \item At submission time, to preserve anonymity, the authors should release anonymized versions (if applicable).
        \item Providing as much information as possible in supplemental material (appended to the paper) is recommended, but including URLs to data and code is permitted.
    \end{itemize}

\item {\bf Experimental setting/details}
    \item[] Question: Does the paper specify all the training and test details (e.g., data splits, hyperparameters, how they were chosen, type of optimizer) necessary to understand the results?
    \item[] Answer: \answerYes{} 
    \item[] Justification: In experimental details part.
    \item[] Guidelines:
    \begin{itemize}
        \item The answer \answerNA{} means that the paper does not include experiments.
        \item The experimental setting should be presented in the core of the paper to a level of detail that is necessary to appreciate the results and make sense of them.
        \item The full details can be provided either with the code, in appendix, or as supplemental material.
    \end{itemize}

\item {\bf Experiment statistical significance}
    \item[] Question: Does the paper report error bars suitably and correctly defined or other appropriate information about the statistical significance of the experiments?
    \item[] Answer: \answerYes{} 
    \item[] Justification: In the main table we discribed the statistical significance.
    \item[] Guidelines:
    \begin{itemize}
        \item The answer \answerNA{} means that the paper does not include experiments.
        \item The authors should answer \answerYes{} if the results are accompanied by error bars, confidence intervals, or statistical significance tests, at least for the experiments that support the main claims of the paper.
        \item The factors of variability that the error bars are capturing should be clearly stated (for example, train/test split, initialization, random drawing of some parameter, or overall run with given experimental conditions).
        \item The method for calculating the error bars should be explained (closed form formula, call to a library function, bootstrap, etc.)
        \item The assumptions made should be given (e.g., Normally distributed errors).
        \item It should be clear whether the error bar is the standard deviation or the standard error of the mean.
        \item It is OK to report 1-sigma error bars, but one should state it. The authors should preferably report a 2-sigma error bar than state that they have a 96\% CI, if the hypothesis of Normality of errors is not verified.
        \item For asymmetric distributions, the authors should be careful not to show in tables or figures symmetric error bars that would yield results that are out of range (e.g., negative error rates).
        \item If error bars are reported in tables or plots, the authors should explain in the text how they were calculated and reference the corresponding figures or tables in the text.
    \end{itemize}

\item {\bf Experiments compute resources}
    \item[] Question: For each experiment, does the paper provide sufficient information on the computer resources (type of compute workers, memory, time of execution) needed to reproduce the experiments?
    \item[] Answer: \answerYes{} 
    \item[] Justification: In the limitations parts we described it.
    \item[] Guidelines:
    \begin{itemize}
        \item The answer \answerNA{} means that the paper does not include experiments.
        \item The paper should indicate the type of compute workers CPU or GPU, internal cluster, or cloud provider, including relevant memory and storage.
        \item The paper should provide the amount of compute required for each of the individual experimental runs as well as estimate the total compute. 
        \item The paper should disclose whether the full research project required more compute than the experiments reported in the paper (e.g., preliminary or failed experiments that didn't make it into the paper). 
    \end{itemize}
    
\item {\bf Code of ethics}
    \item[] Question: Does the research conducted in the paper conform, in every respect, with the NeurIPS Code of Ethics \url{https://neurips.cc/public/EthicsGuidelines}?
    \item[] Answer: \answerYes{} 
    \item[] Justification: 
    \item[] Guidelines:
    \begin{itemize}
        \item The answer \answerNA{} means that the authors have not reviewed the NeurIPS Code of Ethics.
        \item If the authors answer \answerNo, they should explain the special circumstances that require a deviation from the Code of Ethics.
        \item The authors should make sure to preserve anonymity (e.g., if there is a special consideration due to laws or regulations in their jurisdiction).
    \end{itemize}

\item {\bf Broader impacts}
    \item[] Question: Does the paper discuss both potential positive societal impacts and negative societal impacts of the work performed?
    \item[] Answer: \answerYes{} 
    \item[] Justification: In the limitation part we described it.
    \item[] Guidelines:
    \begin{itemize}
        \item The answer \answerNA{} means that there is no societal impact of the work performed.
        \item If the authors answer \answerNA{} or \answerNo, they should explain why their work has no societal impact or why the paper does not address societal impact.
        \item Examples of negative societal impacts include potential malicious or unintended uses (e.g., disinformation, generating fake profiles, surveillance), fairness considerations (e.g., deployment of technologies that could make decisions that unfairly impact specific groups), privacy considerations, and security considerations.
        \item The conference expects that many papers will be foundational research and not tied to particular applications, let alone deployments. However, if there is a direct path to any negative applications, the authors should point it out. For example, it is legitimate to point out that an improvement in the quality of generative models could be used to generate Deepfakes for disinformation. On the other hand, it is not needed to point out that a generic algorithm for optimizing neural networks could enable people to train models that generate Deepfakes faster.
        \item The authors should consider possible harms that could arise when the technology is being used as intended and functioning correctly, harms that could arise when the technology is being used as intended but gives incorrect results, and harms following from (intentional or unintentional) misuse of the technology.
        \item If there are negative societal impacts, the authors could also discuss possible mitigation strategies (e.g., gated release of models, providing defenses in addition to attacks, mechanisms for monitoring misuse, mechanisms to monitor how a system learns from feedback over time, improving the efficiency and accessibility of ML).
    \end{itemize}
    
\item {\bf Safeguards}
    \item[] Question: Does the paper describe safeguards that have been put in place for responsible release of data or models that have a high risk for misuse (e.g., pre-trained language models, image generators, or scraped datasets)?
    \item[] Answer: \answerYes{}
    \item[] Justification: In the limitation part in the Appendix we described.
    \item[] Guidelines:
    \begin{itemize}
        \item The answer \answerNA{} means that the paper poses no such risks.
        \item Released models that have a high risk for misuse or dual-use should be released with necessary safeguards to allow for controlled use of the model, for example by requiring that users adhere to usage guidelines or restrictions to access the model or implementing safety filters. 
        \item Datasets that have been scraped from the Internet could pose safety risks. The authors should describe how they avoided releasing unsafe images.
        \item We recognize that providing effective safeguards is challenging, and many papers do not require this, but we encourage authors to take this into account and make a best faith effort.
    \end{itemize}

\item {\bf Licenses for existing assets}
    \item[] Question: Are the creators or original owners of assets (e.g., code, data, models), used in the paper, properly credited and are the license and terms of use explicitly mentioned and properly respected?
    \item[] Answer: \answerYes{} 
    \item[] Justification: We described them in related works and the main paper.
    \item[] Guidelines:
    \begin{itemize}
        \item The answer \answerNA{} means that the paper does not use existing assets.
        \item The authors should cite the original paper that produced the code package or dataset.
        \item The authors should state which version of the asset is used and, if possible, include a URL.
        \item The name of the license (e.g., CC-BY 4.0) should be included for each asset.
        \item For scraped data from a particular source (e.g., website), the copyright and terms of service of that source should be provided.
        \item If assets are released, the license, copyright information, and terms of use in the package should be provided. For popular datasets, \url{paperswithcode.com/datasets} has curated licenses for some datasets. Their licensing guide can help determine the license of a dataset.
        \item For existing datasets that are re-packaged, both the original license and the license of the derived asset (if it has changed) should be provided.
        \item If this information is not available online, the authors are encouraged to reach out to the asset's creators.
    \end{itemize}

\item {\bf New assets}
    \item[] Question: Are new assets introduced in the paper well documented and is the documentation provided alongside the assets?
    \item[] Answer: \answerNA{} 
    \item[] Justification: Our method does not introduce new assets.
    \item[] Guidelines:
    \begin{itemize}
        \item The answer \answerNA{} means that the paper does not release new assets.
        \item Researchers should communicate the details of the dataset\slash code\slash model as part of their submissions via structured templates. This includes details about training, license, limitations, etc. 
        \item The paper should discuss whether and how consent was obtained from people whose asset is used.
        \item At submission time, remember to anonymize your assets (if applicable). You can either create an anonymized URL or include an anonymized zip file.
    \end{itemize}

\item {\bf Crowdsourcing and research with human subjects}
    \item[] Question: For crowdsourcing experiments and research with human subjects, does the paper include the full text of instructions given to participants and screenshots, if applicable, as well as details about compensation (if any)? 
    \item[] Answer: \answerNA{} 
    \item[] Justification: Our Mu-Edit does not involve crowdsourcing nor research with human subjects.
    \item[] Guidelines:
    \begin{itemize}
        \item The answer \answerNA{} means that the paper does not involve crowdsourcing nor research with human subjects.
        \item Including this information in the supplemental material is fine, but if the main contribution of the paper involves human subjects, then as much detail as possible should be included in the main paper. 
        \item According to the NeurIPS Code of Ethics, workers involved in data collection, curation, or other labor should be paid at least the minimum wage in the country of the data collector. 
    \end{itemize}

\item {\bf Institutional review board (IRB) approvals or equivalent for research with human subjects}
    \item[] Question: Does the paper describe potential risks incurred by study participants, whether such risks were disclosed to the subjects, and whether Institutional Review Board (IRB) approvals (or an equivalent approval/review based on the requirements of your country or institution) were obtained?
    \item[] Answer: \answerNA{} 
    \item[] Justification: Our Mu-Edit does not involve crowdsourcing nor research with human subjects.
    \item[] Guidelines:
    \begin{itemize}
        \item The answer \answerNA{} means that the paper does not involve crowdsourcing nor research with human subjects.
        \item Depending on the country in which research is conducted, IRB approval (or equivalent) may be required for any human subjects research. If you obtained IRB approval, you should clearly state this in the paper. 
        \item We recognize that the procedures for this may vary significantly between institutions and locations, and we expect authors to adhere to the NeurIPS Code of Ethics and the guidelines for their institution. 
        \item For initial submissions, do not include any information that would break anonymity (if applicable), such as the institution conducting the review.
    \end{itemize}

\item {\bf Declaration of LLM usage}
    \item[] Question: Does the paper describe the usage of LLMs if it is an important, original, or non-standard component of the core methods in this research? Note that if the LLM is used only for writing, editing, or formatting purposes and does \emph{not} impact the core methodology, scientific rigor, or originality of the research, declaration is not required.
    \item[] Answer: \answerNo{}
    \item[] Justification: It is only used for polish writing.
    \item[] Guidelines:
    \begin{itemize}
        \item The answer \answerNA{} means that the core method development in this research does not involve LLMs as any important, original, or non-standard components.
        \item Please refer to our LLM policy in the NeurIPS handbook for what should or should not be described.
    \end{itemize}

\end{enumerate}